\documentclass[10pt,journal,compsoc]{IEEEtran}

\usepackage[square,sort,comma,numbers]{natbib}

\usepackage{amsmath}
\usepackage{amsfonts}
\usepackage{amsthm}
\usepackage{amssymb}
\usepackage{mathtools}

\usepackage{background}

\usepackage{multirow}
\usepackage{booktabs}

\usepackage{xspace}

\usepackage{subcaption}
\usepackage{svg}

\usepackage{stfloats}

\usepackage{xcolor}

\usepackage{algorithm}
\usepackage{algorithmic}

\usepackage{tikz}

\usepackage{hyperref}
\usepackage{url}

\usepackage{soul}

\newsavebox{\bigimage}

\usepackage[shortlabels]{enumitem}

\DeclareMathOperator*{\argmax}{arg\,max}
\DeclareMathOperator*{\argmin}{arg\,min}

\newcommand{\cvec}[1]{\boldsymbol{#1}}
\newcommand{\svec}[1]{\mathbf{#1}}
\newcommand{\kldiv}[2]{D_{\text{KL}}\left(#1 \middle\| #2 \right)}
\newcommand{\kldivsmall}[2]{D_{\text{KL}}\left(#1 \middle\| #2 \right)}

\newcommand{\currot}{\textsc{currot}\xspace}
\newcommand{\gradient}{\textsc{gradient}\xspace}
\newcommand{\sprl}{\textsc{sprl}\xspace}

\newcommand{\alpgmm}{\textsc{alp-gmm}\xspace}
\newcommand{\goalgan}{\textsc{goalgan}\xspace}
\newcommand{\acl}{\textsc{acl}\xspace}
\newcommand{\vds}{\textsc{vds}\xspace}
\newcommand{\plr}{\textsc{plr}\xspace}
\newcommand{\her}{\textsc{her}\xspace}

\newcommand{\ppo}{\textsc{ppo}\xspace}
\newcommand{\sac}{\textsc{sac}\xspace}
\newcommand{\dqn}{\textsc{dqn}\xspace}

\newcommand{\rebuttal}[1]{#1}
\newcommand{\revision}[2]{#1}

\definecolor{c2green}{RGB}{81,158,62}
\definecolor{c0blue}{RGB}{58,115,172}

\begin{document}

\title{On the Benefit of Optimal Transport for Curriculum Reinforcement Learning}

\author{Pascal~Klink,
        Carlo~D'Eramo,
        Jan~Peters,
        Joni~Pajarinen
\IEEEcompsocitemizethanks{\IEEEcompsocthanksitem P. Klink and J. Peters are with the Technical University of Darmstadt, Germany, FG Intelligent Autonomous Systems. Correspondence to: pascal@robot-learning.de.
\IEEEcompsocthanksitem J.~Peters is also with the German Research Center for AI (DFKI), Research Department: Systems AI for Robot Learning, hessian.AI (Germany), and the Centre for Cognitive Science at Technical University of Darmstadt.
\IEEEcompsocthanksitem C.~D'Eramo is with the Center for Artificial Intelligence and Data Science at University of W\"{u}rzburg (Germany), the Technical University of Darmstadt (Germany), and hessian.AI (Germany).
\IEEEcompsocthanksitem J.~Pajarinen is with the Department of Electrical Engineering and Automation, Aalto University, Finland. J.~Pajarinen was supported by Academy of Finland (345521).}%
}

\backgroundsetup{contents={\parbox{\textwidth}{\centering This article has been accepted for publication in IEEE Transactions on Pattern Analysis and Machine Intelligence. This is a pre-print version which has not been fully edited and content may change prior to final publication. Citation information: DOI 10.1109/TPAMI.2024.3390051.}},placement=top,scale=0.9,color=black,vshift=-9pt}

\makeatletter               
\AddEverypageHook{%
\SetBgContents{{\parbox{1.1\textwidth}{\centering © 2024 IEEE. Personal use of this material is permitted. Permission from IEEE must be obtained for all other uses, in any current or future media, including reprinting/republishing this material for advertising or promotional purposes, creating new collective works, for resale or redistribution to servers or lists, or reuse of any copyrighted component of this work in other works.}}} 
\SetBgPosition{current page.south}
\SetBgAngle{0}     
\SetBgColor{black}      
\SetBgScale{0.9}    
\SetBgHshift{0}    
\SetBgVshift{43pt}
\bg@material}

\IEEEtitleabstractindextext{%
\begin{abstract}
Curriculum reinforcement learning (CRL) allows solving complex tasks by generating a tailored sequence of learning tasks, starting from easy ones and subsequently increasing their difficulty. Although the potential of curricula in RL has been clearly shown in various works, it is less clear how to generate them for a given learning environment, resulting in various methods aiming to automate this task. In this work, we focus on framing curricula as interpolations between task distributions, which has previously been shown to be a viable approach to CRL. Identifying key issues of existing methods, we frame the generation of a curriculum as a constrained optimal transport problem between task distributions. Benchmarks show that this way of curriculum generation can improve upon existing CRL methods, yielding high performance in various tasks with different characteristics.
\end{abstract}

}

\maketitle

\IEEEdisplaynontitleabstractindextext

\IEEEpeerreviewmaketitle

\IEEEraisesectionheading{\section{Introduction}\label{sec:introduction}}

\IEEEPARstart{R}{einforcement learning} (RL) \citep{sutton1998introduction} has celebrated great successes as a framework for the autonomous acquisition of desired behavior. With ever-increasing computational power, this framework and the algorithms developed under it have resulted in learning agents capable of solving non-trivial long-horizon planning~\citep{mnih2015human,silver2017mastering} and control tasks \citep{akkaya2019solving}. However, these successes have highlighted the need for certain forms of regularization, such as leagues in the context of board games \citep{silver2017mastering}, gradual diversification of simulated training environments for robotic manipulation \citep{akkaya2019solving} and -locomotion \citep{rudin2022learning}, or a tailored training pipeline in the context of humanoid control for soccer \citep{liu2021motor}. These regularizations can help overcome the shortcomings of modern RL agents, such as poor exploratory behavior -- an active research topic \citep{bellemare2016unifying,ghavamzadeh2015bayesian,machado2018count}. 
\\
One can view the regularizations mentioned above under the umbrella term of curriculum reinforcement learning \citep{narvekar2020curriculum}, which aims to avoid the shortcomings of modern (deep) RL agents by learning on a tailored sequence of tasks. Such task sequences can materialize in various ways, and they are motivated by different perspectives in the literature, such as intrinsic motivation or regret minimization, to name some of them \citep{andrychowicz2017hindsight,florensa2017reverse,wang2019poet,portelas2019teacher,wohlke2020performance,jiang2021replay}. \begin{figure}[t!]
    \centering
    \includegraphics{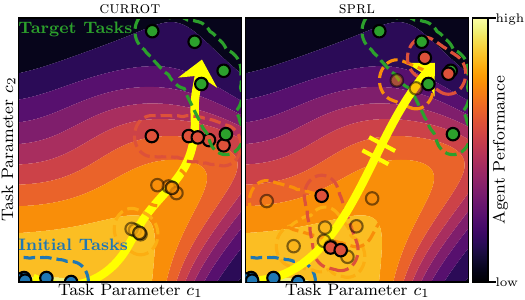}
    \caption{Our approach (\currot) addresses problems of existing curriculum RL methods, such as \sprl, which create curricula between a distribution of initial tasks (blue) and a distribution of target tasks (green). In this example, the curriculum can change the task via two parameters $c_1$ and $c_2$, leading to more or less challenging learning environments for an agent. Looking at the different stages of the curricula (colored points), we see that existing methods can lead to distributions that encode hard- and easy tasks, but ignore tasks of intermediate difficulty. Our method avoids such a splitting behavior, resulting in interpolations that gradually increase the task difficulty throughout the curriculum. Please see Sections \ref{sec:currot:crlasot} and \ref{sec:currot:approx_currot} for a detailed description.}
    \label{fig:currot:pull_figure}
    \vspace{-8pt}
\end{figure} \\
A perspective of particular interest for this article is to interpret a curriculum as a sequence of task distributions that interpolate between an auxiliary task distribution -- with the sole purpose of facilitating learning -- and a distribution of target tasks \citep{klink2021probabilistic}. We refer to these approaches as interpolation-based curricula. While algorithmic realizations of such curricula have been successfully evaluated in the literature \cite{klink2019self,klink2020self,chen2021variational}, some evaluations indicated a relatively poor learning performance of these methods \citep{romac2021teachmyagent}. Furthermore, applications of interpolation-based curricula have been limited to scenarios with somewhat restricted distributions, such as Gaussian- or uniform ones. The observed performance gaps and lack of flexibility w.r.t. distribution parameterization call for a better understanding of these methods' inner workings to improve their performance and extend their applicability. \\
This article investigates the shortcomings of methods that realize curricula as a scheduled interpolation between task distributions based on the KL divergence and an expected performance constraint. We show how both these concepts can fail to produce meaningful curricula in simple examples. \\
The demonstrated failure cases a) illustrate the importance of explicitly reasoning about the similarity of tasks when building a curriculum and b) show how parametric assumptions on the generated task distributions can masquerade failures of the underlying framework used to generate curricula. To resolve the observed issues, we explicitly specify the similarity of learning tasks via a distance function and use the framework of optimal transport to generate interpolating distributions that, independent of their parameterization, result in gradual task changes. Based on this explicit notion of task similarity, we propose our approach to curriculum RL (\currot), which replaces the expected performance constraint with a more strict condition to obtain the behavior visualized in Figure~\ref{fig:currot:pull_figure}. Furthermore, we contrast our approach with an alternative method, \gradient, recently proposed by \citet{huang2022curriculum}. We outline how both approaches use optimal transport to generate curricula but differ in their use of the agent performance to constrain the curriculum while avoiding the demonstrated pitfalls of expected performance constraints. \\
In experiments, we a) validate the correct behavior of both \currot and \gradient free from approximations and parametric assumptions in a small discrete MDP and b) compare approximate implementations on a variety of tasks featuring discrete- and continuous task spaces, as well as Euclidean- and non-Euclidean measures of distance between learning tasks. In these experiments, both approaches show convincing performance with \currot consistently matching and surpassing the performance of all other algorithms.

\section{Related Work}

This work generates training curricula for reinforcement learning (RL) agents. Unlike supervised learning, where there is an ongoing discussion about the mechanics and effects of curricula in different learning situations \citep{weinshall2020theory,wu2021when}, the mechanics seem to be more agreed upon in RL. \\
\textbf{Curriculum Reinforcement Learning:} In RL, curricula improve the learning performance of an agent by adapting the training environments to its proficiency\rebuttal{. This adaptation of task complexity can reduce the sample complexity of RL, e.g., by bypassing poor exploratory behavior of non-proficient agents \cite{li2023understanding}.} Using curricula can avoid the need for extensively engineered reward functions, which come with risks, such as failing to encode the intended behavior \cite{allievi2023perils}. Applications of curricula to RL are widespread, and different terms have been established. Adaptive Domain Randomization \citep{akkaya2019solving} uses curricula to gradually diversify the training parameters of a simulator to facilitate sim-to-real transfer. Similarly, unsupervised environment discovery \citep{dennis2020emergent,jiang2021prioritized,jiang2021replay} aims to efficiently train an agent robust to variations in the environment and can be seen as a more general view of domain randomization. Automatic curriculum learning methods \citep{florensa2017reverse,sukhbaatar2018intrinsic,florensa2018automatic,portelas2019teacher,zhang2020automatic,racaniere2020automated,eimer2021self,klink2021probabilistic} mainly focus on improving an agent's learning speed or performance on a set of desired tasks. Curricula are often generated as distributions that maximize a specific surrogate objective, such as learning progress \citep{baranes2010intrinsically,portelas2019teacher}, intermediate task difficulty \citep{florensa2018automatic}, regret \citep{jiang2021prioritized}, or disagreement between $Q$-functions \citep{zhang2020automatic}. Curriculum generation can also be interpreted as a two-player game \citep{sukhbaatar2018intrinsic}. The work by \citet{jiang2021replay} hints at a link between surrogate objectives and two-player games. Similar to the variety of objectives that the above algorithms optimize to build a curriculum, their implementations use drastically different approaches to approximate the training distribution for the agent, which is often defined over a continuous space of training tasks. For example, \citet{florensa2018automatic} use a combination of GANs and a replay buffer to represent the task distribution. \citet{portelas2019teacher} use a Gaussian mixture model to approximate the distribution of tasks that promise high learning progress. \citet{jiang2021replay} use a fixed-size replay buffer to realize an approximate distribution of high-regret tasks, simultaneously encouraging frequent replay of buffered tasks to keep a more accurate estimate of regret. \\
Interpolation-based curriculum RL algorithms formulate the generation of a curriculum as an explicit interpolation between an auxiliary task distribution and a distribution of target tasks \citep{klink2019self,klink2021probabilistic,chen2021variational}. This interpolation is subject to a constraint on the expected agent performance that paces its progress toward the target tasks. As highlighted by \citet{klink2021probabilistic}, such interpolations can be formally linked to successful curricula in supervised learning \citep{kumar2010self}, the concept of annealing in statistics \citep{neal2001annealed}, and homotopic continuation methods in optimization \citep{allgower2003introduction}. As for the algorithms based on surrogate objectives, realizations of these interpolation-based curricula inevitably need to rely on approximations such as the restriction to Gaussian distributions in \cite{klink2019self,klink2020self,klink2021probabilistic} or approximate update rules enabled by uniform target task distributions \cite{chen2021variational}. \\
This article reveals shortcomings of the aforementioned interpolation-based curriculum RL methods, highlighting how approximations can masquerade issues in the conceptual algorithm formulations. One ingredient to overcome these shortcomings is an explicit notion of task similarity that we formulate as a distance function between tasks. We can then lift this distance function into the space of probability measures using \emph{optimal transport}. \\
\textbf{Optimal Transport:} Dating back to work by Monge in the 18th century, \emph{optimal transport} has been understood as a fundamental concept touching upon many fields in both theory and application \citep{peyre2019computational,chen2021stochastic}. In probability theory, optimal transport translates to the so-called Wasserstein metric \citep{kantorovich1942transfer} that compares two distributions under a given metric, allowing, e.g., for the analysis of probabilistic inference algorithms as approximate gradient flows \citep{liu2018understanding} and providing well-defined ways of comparing feature distributions or even graphs in computer vision and machine learning \cite{kolouri2017optimal,kandasamy2018neural,togninalli2019wasserstein}. Gromov-Wasserstein distances \cite{memoli2011gromov,vincent-cuaz2022semirelaxed} even allow comparing distributions across metric spaces, which has been of use, e.g., in computational biology \cite{demetci2020gromov} or imitation learning \cite{fickinger2022crossdomain}. In Reinforcement learning, optimal transport has not found widespread application, albeit some interesting works exist. \citet{zhang2018policy} provide a natural extension of the work by \citet{liu2018understanding} and interpret policy optimization as Wasserstein gradient flows. \citet{metelli2019propagating} use Wasserstein barycenters to propagate uncertainty about value function estimates in a $Q$-learning approach. In more applied scenarios, optimal transport has been used to regularize RL in sequence generation- \citep{chen2020sequence} or combinatorial optimization problems \citep{goh2022combining}. \rebuttal{In goal-conditioned RL, Wasserstein distances have been previously applied to improve goal generation in the hindsight experience replay framework \cite{ren2019exploration} and to realize well-performing data-driven reward functions by combining them with so-called time-step metrics \citep{durugkar2021adversarial}. Recently, \citet{cho2023outcome} combined the data-driven reward function proposed by \citet{durugkar2021adversarial} with a curriculum that, similarly to the work by \cite{ren2019exploration}, improves the selection of training goals from a buffer of achieved ones.} When it comes to building \rebuttal{RL curricula over arbitrary MDPs using Optimal Transport}, we are only aware of our work \citep{klink2022curriculum} at ICML 2022 and the work by \cite{huang2022curriculum} at NeurIPS 2022, which we present from a unified perspective and compare in this journal article\rebuttal{. In addition to the aforementioned methods in goal-conditioned RL, this article emphasizes curriculum reinforcement learning as another promising application domain for optimal transport.} An important issue of applied optimal transport is its computational complexity. In Appendix \ref{app:currot:ot-complexity}, we discuss the computational aspects of optimal transport in more detail.

\section{Preliminaries}

This section introduces the necessary background on (contextual) RL, curriculum RL, and optimal transport.

\subsection{Contextual Reinforcement Learning}

Contextual reinforcement learning \citep{hallak2015contextual} can be seen as a conceptual extension to the (single task) reinforcement learning (RL) problem
\begin{align}
    \max_{\pi} J(\pi) &= \max_{\pi} \mathbb{E}_{p(\cvec{\tau} \vert \pi)} \left[\sum_{t=0}^{\infty} \gamma^t r(\svec{s}_t, \svec{a}_t) \right] \label{eq:currot:rl-objective} \\
    \cvec{\tau} &= \left\{(\svec{s}_t, \svec{a}_t) \middle| t=1,\ldots \right\} \nonumber \\
    p(\cvec{\tau} \vert \pi) &= p_0(\svec{s}_0) \prod_{t=1}^{\infty} p(\svec{s}_{t} \vert \svec{s}_{t-1}, \svec{a}_{t-1}) \pi(\svec{a}_{t-1} \vert \svec{s}_{t-1}), \nonumber
\end{align}
which aims to maximize the above expected discounted reward objective by finding an optimal policy ${\pi{:} \mathcal{S} {\times} \mathcal{A} \mapsto \mathbb{R}_{\geq 0}}$ for a given MDP $\mathcal{M} {=} \langle \mathcal{S}, \mathcal{A}, p, r, p_0 \rangle$ with initial state distribution $p_0$ and transition dynamics $p$. Contextual RL extends this objective to a space of MDPs ${\mathcal{M}(\svec{c}) {=} \langle \mathcal{S}, \mathcal{A}, p_{\svec{c}}, r_{\svec{c}}, p_{0, \svec{c}} \rangle}$ equipped with a distribution $\mu{:} \mathcal{C} {\mapsto} \mathbb{R}$ over contextual variables $\svec{c} \in \mathcal{C}$
\begin{align}
       \max_{\pi} J(\pi, \mu) = \max_{\pi} \mathbb{E}_{\mu(\svec{c})} \left[ J(\pi, \svec{c}) \right]. \label{eq:currot:con-rl-objective}
\end{align}
The policy $\pi: \mathcal{S} {\times} \mathcal{C} {\times} \mathcal{A} \mapsto \mathbb{R}$ is conditioned on the contextual parameter $\svec{c}$. The distribution $\mu(\svec{c})$ encodes the tasks $\mathcal{M}({\svec{c}})$ to be solved by the agent. Objective $J(\pi, \svec{c})$ in Eq. (\ref{eq:currot:con-rl-objective}) corresponds to objective $J(\pi)$ in Eq. (\ref{eq:currot:rl-objective}) with the initial state distribution $p_{0}$, the transition dynamics $p$ as well as the reward function $r$ of $\mathcal{M}$ replaced by their counterparts in $\mathcal{M}(\svec{c})$. This contextual model of optimal decision-making is well-suited for learning in multiple related tasks, as is the case in multi-task- \citep{wilson2007multi}, goal-conditioned- \citep{schaul2015universal}, or curriculum RL \citep{narvekar2020curriculum}. At this point, we want to emphasize that the context~$\svec{c}$ could be readily embedded in the state space $\mathcal{S}$, resulting in a regular MDP in which the context -- as part of the state -- remains constant throughout an episode. The context distribution $\mu(\svec{c})$ would then be subsumed into the initial state distribution without losing expressiveness. We nonetheless prefer the contextual RL framework, coined by \citet{hallak2015contextual}, as it emphasizes the distribution $\mu(\svec{c})$, which is at the heart of curriculum RL methods, as we will see now.

\subsection{Curriculum Reinforcement Learning}

On an abstract level, curriculum RL methods can be understood as generating a sequence of task distributions $\left(p_i{:} \mathcal{C} {\mapsto} \mathbb{R}\right)_{i}$ under which to train an RL agent by maximizing $J(\pi, p_i)$ w.r.t. $\pi$. When chosen appropriately, solving this sequence of optimization problems can yield a policy that performs better on the target distribution $\mu(\svec{c})$ than a policy found by maximizing $J(\pi, \mu)$ directly. The benefit of such mediating distributions is particularly obvious in settings where initially random agent behavior is unlikely to observe any meaningful learning signals, such as in sparse-reward learning tasks. \\
CRL methods differ in the specification of $p_i$. Often, the distribution is defined to prioritize tasks that maximize certain surrogate quantities, such as absolute learning progress \citep{portelas2019teacher}, regret \citep{jiang2021prioritized}, or tasks of intermediate success probability \citep{florensa2018automatic}. This article focuses on CRL methods that model $p_i$ as the solution to an optimization problem that aims to minimize a distance or divergence between $p_i$ and $\mu$. One of these approaches \citep{klink2019self,klink2020self,klink2021probabilistic} defines $p_i$ as the distribution with minimum KL divergence to $\mu$ that fulfills a constraint on the expected agent performance
\begin{align}
	\min_{p}&\ \kldiv{p(\svec{c})}{\mu(\svec{c})} \label{eq:currot:sprl-objective} \\  
	\text{s.t.}& \ \ J(\pi, p) \geq \delta \qquad \kldiv{p(\svec{c})}{q(\svec{c})} \leq \epsilon, \nonumber
\end{align}
where $\delta$ is the desired level of performance to be achieved by the agent $\pi$ under $p(\svec{c})$ and $\epsilon$ limits the maximum KL divergence to the previous context distribution $q(\svec{c}) {=} p_{i-1}(\svec{c})$. The optimizer of (\ref{eq:currot:sprl-objective}) balances between tasks likely under the (target) distribution $\mu(\svec{c})$ and tasks in which the agent currently obtains large rewards. The KL divergence constraint w.r.t. the previous context distribution $q(\svec{c})$ prevents large changes in $p(\svec{c})$ during subsequent iterations, avoiding the exploitation of faulty estimates of the agent performance $J(\pi, p)$ from a limited amount of samples. Objective (\ref{eq:currot:sprl-objective}) performs an interpolation between the distributions $p_{\eta}(\svec{c}) {\propto} \mu(\svec{c}) \exp(\eta J(\pi, \svec{c}))$ and $q(\svec{c})$, given by
\begin{align}
    p_{\alpha, \eta}(\svec{c}) \propto \left(\mu(\svec{c}) \exp(J(\pi, \svec{c}))^{\eta} \right)^{\alpha} q(\svec{c})^{1 - \alpha}. \label{eq:currot:analytic-dist}
\end{align}
The two parameters $\alpha$ and $\eta$ that control the interpolation are the Lagrangian multipliers of the two constraints in objective (\ref{eq:currot:sprl-objective}). We will later investigate the behavior of this interpolating distribution.

\subsection{Optimal Transport}
\label{sec:currot:ot}

The problem of optimally transporting density between two distributions has been initially investigated by \citet{monge1781memoire}. As of today, generalizations established by \citet{kantorovich1942transfer} have led to so-called \textbf{Wasserstein distances} as metrics between probability distributions defined on a metric space $M{=}(d, \mathcal{C})$ with metric $d: \mathcal{C} \times \mathcal{C} \mapsto \mathbb{R}_{\geq 0}$
\begin{align*}
\mathcal{W}_p(p_1, p_2) &{=} \left( \inf_{\phi \in \Phi(p_1, p_2)}  \mathbb{E}_{\phi} \left[d(\svec{c}_1, \svec{c}_2)^p \right] \right)^{1 / p}, \quad p \geq 1 \\ 
\Phi(p_1, p_2) &{=} \left\{ \phi: \mathcal{C} {\times} \mathcal{C} {\mapsto} \mathbb{R}_{\geq 0} \middle| p_i {=} \rebuttal{P_{i\#}} \phi,\ i {\in} \{1, 2\} \right\},
\end{align*}
\rebuttal{where $P_{i\#}$ are the push-forwards of the maps ${P_1(\svec{c}_1, \svec{c}_2) {=} \svec{c}_1}$ and $P_2(\svec{c}_1, \svec{c}_2) {=} \svec{c}_2 $. We refer to \cite[Chapter 2]{peyre2019computational} for an excellent and intuitive introduction to these concepts.} The distance between $p_1$ and $p_2$ is obtained via the solution to an optimization problem that finds a so-called plan\rebuttal{, or coupling,} $\phi$. This \rebuttal{coupling} encodes how to equalize $p_1$ and $p_2$, considering the cost of moving density between parts of the space $\mathcal{C}$. The metric $d$ encodes this cost. In the following, we will always assume to work with $2$-Wasserstein distances, i.e., $p{=}2$, due to their suitedness for interpolating measures \cite[see][Chapter 6 and Remark 2.24]{peyre2019computational}. \\
Similar to how (weighted) means can be defined as solutions to optimization problems on a metric space $M{=}(d, \mathcal{C})$, Wasserstein distances allow us to define what is referred to as Wasserstein barycenters \citep{agueh2011barycenters}
\begin{align}
    \mathcal{B}_2(W, P) = \argmin_{p} \sum_{k=1}^K w_k \mathcal{W}_2(p, p_k), \label{eq:currot:wasserstein-barycenters}
\end{align}
which represent the (weighted) mean of the distributions $P {=} \{p_k | k {\in} [1,K] \}$ with weights $W {=} \{w_k | k {\in} [1,K] \}$.

\section{Curriculum Reinforcement Learning as Constrained Optimal Transport}
\label{sec:currot:crlasot}

At this point, we can motivate our approach to curriculum RL by looking at the limitations of Objective \ref{eq:currot:sprl-objective} caused by a) measuring similarity between context distributions via the KL divergence and b) the expected performance constraint used to control the progression towards $\mu(\svec{c})$.

\subsection{Limitations of the KL Divergence}

\begin{figure}[b!]
    \centering
    \includegraphics{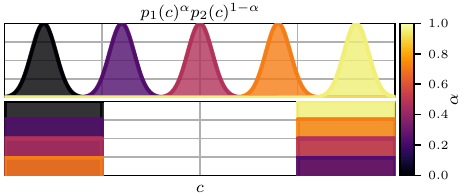}
    \vspace{-5pt}
    \caption{Interpolations generated by optimizing Objective (\ref{eq:currot:kl-interp}) for different values of $\epsilon$ (and with that $\alpha$). In the top row, $p_1(c)$ and $p_2(c)$ are Gaussian, while in the bottom row, they assign uniform density over different parts of $\mathcal{C}$.}
    \label{fig:currot:kl_interpolations}
\end{figure}

\begin{figure}[t]
    \centering
    \includegraphics{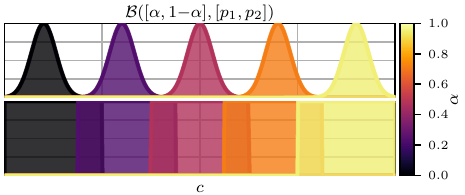}
    \vspace{-5pt}
    \caption{Wasserstein barycenters $\mathcal{B}([\alpha, 1{-}\alpha], [p_1, p_2])$ between the distributions shown in Figure \ref{fig:currot:kl_interpolations}. In the top row, $p_1(c)$ and $p_2(c)$ are Gaussian while in the bottom row, they assign uniform density over different parts of $\mathcal{C}$.}
    \label{fig:currot:wasserstein_interpolations}
\end{figure}

Given the complexity of computing $\kldiv{p(\svec{c})}{\mu(\svec{c})}$ for arbitrary distributions, previous work restricts $\mu(\svec{c})$ either to a Gaussian distribution \citep{klink2019self,klink2020self,klink2021probabilistic} or to be uniform over $\mathcal{C}$ to ease computation and optimization of a weighted KL divergence objective \citep{chen2021variational}. While empirically successful, these design choices masquerade the pitfalls of the KL divergence to measure distributional similarity in a CRL setting, particularly when dealing with a target distribution that does not assign uniform density over all of $\mathcal{C}$. \\
To demonstrate this issue, we focus on an interpolation task between two distributions 
\begin{align}
    p_1(\svec{c})^{\alpha(\epsilon)} p_2(\svec{c})^{1 - \alpha(\epsilon)} &= \argmin_{p \in \{ q \vert \kldivsmall{q}{p_2} \leq \epsilon \}} \kldiv{p}{p_1},\label{eq:currot:kl-interp}
\end{align}
corresponding to a version of Objective (\ref{eq:currot:sprl-objective}) with no constraint on the expected agent performance. Figure \ref{fig:currot:kl_interpolations} demonstrates the sensibility of this interpolation to the parametric representation of the distributions $\mu(\svec{c})$ and $q(\svec{c})$. While for Gaussian distributions, interpolations of the form $p_1(\svec{c})^{\alpha} p_2(\svec{c})^{1 - \alpha}$ gradually shift density in a metric sense, this behavior is all but guaranteed for non-Gaussian distributions. The interpolation between two uniform distributions with quasi-limited support \footnote{We ensure a negligible positive probability density across all of $\mathcal{C}$ to allow for the computation of KL divergences.} in the bottom row of Figure \ref{fig:currot:kl_interpolations} displaces density from contexts $\svec{c}$ to contexts $\svec{c}'$ with large Euclidean distance $\| \svec{c} - \svec{c}' \|_2$. In settings in which the Euclidean distance between contexts $\svec{c}_1$ and $\svec{c}_2$ is a good indicator for the similarity between $\mathcal{M}(\svec{c}_1)$ and $\mathcal{M}(\svec{c}_2)$, the observed ignorance of the KL divergence w.r.t. the underlying geometry of the context space leads to curricula with ``jumps'' in task similarity. We can easily convince ourselves that such jumps are not a hypothetical problem by recalling that neural network-based policies $\pi(\svec{a} | \svec{s}, \svec{c}) {=} \svec{f}_{\boldsymbol{\theta}}(\svec{s}, \svec{c})$ tend to gradually change their behavior with increasing Euclidean distance to $\svec{c}$. \\
At this point, we can leverage the notion of optimal transport to explicitly encode the similarity of two tasks, $\mathcal{M}(\svec{c})$ and $\mathcal{M}(\svec{c}')$, via a metric $d(\svec{c}, \svec{c}')$ and realize the interpolation between distributions on the resulting metric space as Wasserstein barycenters (Eq. \ref{eq:currot:wasserstein-barycenters}). As we see in Figure \ref{fig:currot:wasserstein_interpolations}, this explicit notion of task similarity allows to generate interpolations that are stable across changes in the parameterization of context distributions and interpolate between arbitrary distributions that are not absolutely continuous w.r.t. each other. Consequently, the optimization problem
\begin{align}
    \min_{p} \mathcal{W}_2(p, \mu)\;\, \text{s.t.}\ J(\pi, p) \geq \delta \label{eq:currot:flawed-version}
\end{align}
is a promising approach to leverage optimal transport in curriculum RL.
We iterate on this candidate in the next section by investigating the role of the expected performance constraint when generating curricula for reinforcement learning agents.

\subsection{Challenges of Expected Performance Constraints}

\begin{figure}[t]
    \centering
    \includegraphics{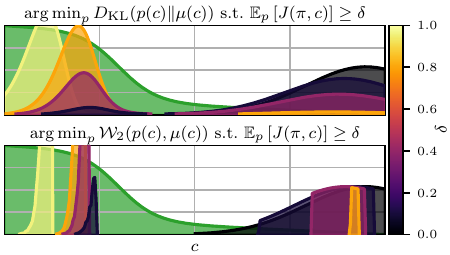}
    \vspace{-5pt}
    \caption{Interpolations using KL divergence (top) and Wasserstein distance (bottom) subject to an expected performance constraint with different threshold values $\delta$. The performance $J(\pi, c)$ is visualized in green.}
    \label{fig:currot:ep_interpolations}
\end{figure}

\begin{figure}[t]
    \centering
    \includegraphics{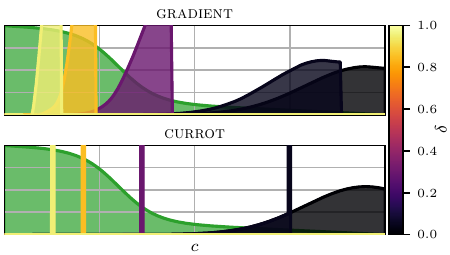}
    \vspace{-5pt}
    \caption{Interpolations generated by \gradient (Eq. \ref{eq:currot:gradient}, top) and \currot (Eq. \ref{eq:currot:currot}, bottom) for different threshold values $\delta$. The performance $J(\pi, c)$ is visualized in green.}
    \label{fig:currot:currot_gradient_interpolations}
\end{figure}

The \sprl objective (\ref{eq:currot:sprl-objective}) controls the interpolation speed between the initial- and target task distribution by the expected performance of the current agent under the chosen context distribution $J(\pi, p)$. As detailed in \citep{klink2021probabilistic}, this expected performance constraint allows for establishing a connection to self-paced learning for supervised learning tasks \cite{kumar2010self,meng2017theoretical}. While this formal connection is interesting in its own right, we show in Figure \ref{fig:currot:ep_interpolations} that the expected performance constraint in \sprl can lead to encoding both too simple and too complex tasks, given the current agent capabilities. Furthermore, using Wasserstein distances in Objective (\ref{eq:currot:flawed-version}) does not resolve this issue. \rebuttal{In Figure \ref{fig:currot:ep_interpolations}, both methods encode tasks with very high and very low agent returns to fulfill the expected performance constraint, sidestepping the goal of encoding tasks of intermediate difficulty.}  At this point, we can propose our algorithm \currot and introduce a recent algorithm proposed by \citet{huang2022curriculum} -- called \gradient -- as two ways of resolving the observed interpolation issue:
\begin{enumerate}
    \item \textbf{\currot} restricts the support of $p(\svec{c})$ to those contexts $\svec{c} \in \mathcal{C}$ that fulfill the performance constraint $J(\pi, \svec{c}) {\geq} \delta$. We refer to this set as $\mathcal{V}(\pi, \delta) = \{\svec{c} | \svec{c} {\in} \mathcal{C}, J(\pi, \svec{c}) \geq \delta\}$. With this notation in place, we frame the restricted optimization as
    \begin{align}
        \min_{p} \mathcal{W}_2(p, \mu) \quad \text{s.t.} \;\, p(\mathcal{V}(\pi, \delta)) {=} 1. \label{eq:currot:currot}
    \end{align}
    Putting the constraint in words,  we require that the curriculum assigns all probability density of $p$ to contexts that satisfy the performance constraint. 
    \item \textbf{\gradient} restricts the interpolation to the barycentric interpolation (\ref{eq:currot:wasserstein-barycenters}) between the initial- and target context distribution, i.e. $p_{\alpha}(\svec{c}) = \mathcal{B}_2\left( [1{-}\alpha, \alpha], [p_0(\svec{c}), \mu(\svec{c})] \right)$. \rebuttal{This restriction prevents the problematic behavior shown in Figure \ref{fig:currot:ep_interpolations}} while still allowing to adjust $\alpha$ using an expected performance constraint
    \begin{align}
        \max_{\alpha \in [0, 1]} \alpha \;\, \text{s.t.} \ J(\pi, p_{\alpha}) {\geq} \delta. \label{eq:currot:gradient} \end{align}
\end{enumerate}
As shown in Figure \ref{fig:currot:currot_gradient_interpolations}, both of these methods avoid the behavior generated by Objective (\ref{eq:currot:flawed-version}), resulting in an interpolation that gradually deforms the distribution in a metric sense with changing agent competence. In the remainder of this article, we will investigate exact and approximate versions of these algorithms to understand their behavior better. The first observation in this regard is that the curriculum of \gradient is entirely predetermined by the given metric $d(\svec{c}_1, \svec{c}_2)$ as well as the target- and initial distribution $\mu(\svec{c})$ and $p_0(\svec{c})$. The agent performance only influences how fast the curriculum proceeds towards $\mu(\svec{c})$. On the other hand, \currot reshapes the curriculum based on the current agent performance to avoid sampling contexts with a performance lower than the threshold $\delta$. Figure \ref{fig:currot:currot_gradient_interpolations} shows that this reshaping results in a tendency of \currot to place all probability density on the border of the desired agent performance $\delta$ until reaching regions of non-zero probability density under $\mu(\svec{c})$. At this point, the curriculum matches the target density in those parts of $\mathcal{C}$, in which the performance constraint is fulfilled, and continues to concentrate all remaining density on the boundaries of agent capability. This behavior is similar to those CRL methods that combine task-prioritization with a replay buffer of, e.g., previously solved tasks to prevent catastrophic forgetting, such as \goalgan or \plr \citep{florensa2018automatic,jiang2021prioritized}. To the best of our knowledge, such behavior has not yet been motivated by a first-principle optimization objective in the context of curriculum RL.

\section{Approximate Algorithms for Discrete- and Continuous Context Spaces}
\label{sec:currot:approx_currot}

Objectives (\ref{eq:currot:currot}) and (\ref{eq:currot:gradient}) face challenges in more realistic application scenarios with either large discrete- or continuous context spaces due to two reasons:
\begin{enumerate}
    \item We do not have access to the expected performance $J(\pi, \svec{c})$ of an agent $\pi$ in context $\svec{c}$ but can only estimate it from observed training episodes.
    \item Computing Wasserstein barycenters for arbitrary continuous- or discrete distributions in non-Euclidean spaces can quickly become intractably expensive.
\end{enumerate}
The following sections address the above problems to benchmark \currot and \gradient in non-trivial experimental settings.

\subsection{\rebuttal{Approximate Wasserstein Barycenters}}

\rebuttal{Before branching into the description of the two algorithms, we first describe a particle-based approximation to the computation of Wasserstein Barycenters, which allows us to cheaply approximate Barycenters for the \gradient algorithm in large discrete state-spaces and is essential for the approximate implementation of the \currot algorithm. \\
For approximating a Barycenter $p_{\alpha} {=} \mathcal{B}([1 - \alpha, \alpha], [p_0, \mu])$, we first sample a set of $N$ particles from $\mu(\svec{c})$ and $p_0(\svec{c})$ to form the empirical distributions
\begin{align}
    \hat{\mu}(\svec{c}) &= \frac{1}{N} \sum_{n=1}^N \delta_{\svec{c}_{\mu,n}}(\svec{c}),\quad \svec{c}_{\mu,n} \sim \mu(\svec{c}) \label{eq:currot:empirical-dist} \\
    \hat{p}_0(\svec{c}) &= \frac{1}{N} \sum_{n=1}^N \delta_{\svec{c}_{p_0,n}}(\svec{c}),\quad \svec{c}_{p_0,n} \sim p_0(\svec{c}), \nonumber
\end{align}
where $\delta_{\svec{c}_{\text{ref}}}(\svec{c})$ represents a Dirac distribution centered at $\svec{c}_{\text{ref}}$. Due to the discrete nature of $\hat{\mu}(\svec{c})$ and $\hat{p}_0(\svec{c})$, the coupling $\phi(\svec{c}_1, \svec{c}_2)$ reduces to a permutation $\phi {\in} \text{Perm}(N)$, which assigns the particles between $\hat{p}_0$ and $\hat{\mu}$ \cite[Section 2.3]{peyre2019computational}. With that, the computation of $\mathcal{W}_2(\hat{p}_0, \hat{\mu})$ reduces to
\begin{align}
\min_{\rebuttal{\phi \in \text{Perm}(N)}} \left(\frac{1}{N} \sum_{n=1}^N d(\svec{c}_{p_0,n}, \svec{c}_{\mu,\rebuttal{\phi(n)}})^2\right)^{\frac{1}{2}}. \label{eq:currot:assigment}
\end{align}
Since a permutation is a particular case of a coupling \cite[Section 2.3]{peyre2019computational}, we overload the meaning of $\phi$ to be either a permutation or coupling, depending on the number of arguments. With today's computing hardware, assignment problems like (\ref{eq:currot:assigment}) can be solved on a single CPU core in less than a second for $N$ in the hundreds, which is typically enough to represent the context distributions~\footnote{In our experiments, we use less than a thousand particles in all experiments}. Given this optimal assignment, we then compute the Fréchet mean for each particle pair
\begin{align}
    \svec{c}_{\alpha,n} {=} \argmin_{\svec{c} \in \mathcal{C}} (1 {-} \alpha) d(\svec{c}, \svec{c}_{p_0,n})^2 + \alpha d(\svec{c}, \svec{c}_{\mu,\rebuttal{\phi(n)}})^2 \label{eq:currot:frechet}
\end{align}
to form the barycenter $\hat{p}_{\alpha}(\svec{c}) {=} \frac{1}{N} \sum_{n=1}^N \delta_{\svec{c}_{\alpha,n}}(\svec{c})$. While certainly less efficient than specialized routines for Barycenter computations in Euclidean Spaces, such as e.g., the GeomLoss library \cite{feydy2019interpolating}, the presented approach is useful when dealing with large discrete spaces. In this case, faithful Barycenter computations must work with the full distance matrix. Assuming a discrete context space of size $S$ and neglecting the cost of computing the optimal assignment, the approximate barycenter computation requires $O(N^2 + 2NS)$ evaluations of the distance function. Hence for $S \gg N$, even computing the $\frac{S(S+1)}{2}$ entries of the entire distance matrix required for a single step in the Sinkhorn algorithm becomes more expensive than the presented approximate method. Additionally, reducing the Barycenter computation to an optimization problem over individual particles easily allows to incorporate additional constraints that are required by the \currot optimization objective (\ref{eq:currot:currot}).}

\subsection{Approximate \gradient}

\begin{algorithm}[t]
\caption{Approximate \gradient}\label{alg:currot:gradient}
\begin{algorithmic}
\STATE {\bfseries Input:} Initial context dist.\ $p_0(\svec{c})$, target context dist.\ $\mu(\svec{c})$, metric $d(\svec{c}_1, \svec{c}_2)$, performance bound $\delta$, step size $\epsilon$
    \STATE {\textbf{Initialize:} $\alpha = 0$}
    \WHILE{True}
    \STATE {Compute $\hat{p}_{\alpha}(\svec{c}) = \frac{1}{N} \sum_{n=1}^N \delta_{\svec{c}_{\alpha, n}}(\svec{c})$ (Eq. (\ref{eq:currot:assigment}) and (\ref{eq:currot:frechet}))}
    \STATE {\bfseries Agent Improvement:}
   \STATE {Sample contexts $\svec{c}_m \sim \hat{p}_{\alpha}(\svec{c}),\ m \in [1, M]$}
   \STATE {Train policy $\pi$ under $\svec{c}_m$ and observe episodic rewards $R_m = \sum_{t=0}^\infty \gamma^t r_{\svec{c}_m}(\svec{s}_t, \svec{a}_t),\ m \in [1, M]$}
   \STATE {\bfseries Context Distribution Update:}
   \IF{$\frac{1}{M} \sum_{m=1}^M R_m {\geq} \delta$}
        \STATE {Advance interpolation $\alpha = \text{min}(\alpha + \epsilon, 1)$}
   \ENDIF
   \ENDWHILE
\end{algorithmic}
\end{algorithm}

\noindent \citet{huang2022curriculum} propose to compute barycenters between $p_0(\svec{c})$ and $\mu(\svec{c})$ for discrete steps of size $\epsilon$. Starting from $\alpha{=}0$, the agent trains for $M$ episodes on tasks sampled from the current distribution. If the average episodic return $\frac{1}{M} \sum_{m=1}^M R_m$ is greater or equal to $\delta$, $\alpha$ is increased by $\epsilon$ and the distribution is set to be the Wasserstein barycenter for the updated value of $\alpha$. \\
This step-wise increase of $\alpha$ avoids the explicit optimization over $\alpha$ and, with that, the need to estimate the performance of the current policy $\pi$ for a given context $\svec{c}$. \rebuttal{Having laid out a way of computing approximate Barycenters in the previous section, we can summarize our implementation of \gradient in Algorithm \ref{alg:currot:gradient}.}

\subsection{Approximate \currot}

\noindent As for the \gradient algorithm, we make use of an empirical distribution $\hat{p}(\svec{c})$ to represent the context distribution $p(\svec{c})$ (see Eq. \ref{eq:currot:empirical-dist}). Unlike for \gradient, there is no possibility to side-step the estimation of $J(\pi, \svec{c})$ for \currot, and any estimator of $J(\pi, \svec{c})$ will inevitably make mistakes. The mistakes will be particularly big for contexts $\svec{c}$ with a considerable distance to those sampled under the current training distribution $p(\svec{c})$. To avoid exploiting such erroneous performance predictions, we introduce a trust region constraint similar to the seminal \sprl objective (\ref{eq:currot:sprl-objective}) into \currot
\begin{align}
    \min_{p}&\ \mathcal{W}_2(p, \mu) \quad \label{eq:currot:approximate-currot} \\
    \text{s.t.}& \ \ p(\mathcal{V}(\pi, \delta)) {=} 1 \qquad \mathcal{W}_2(p, q) \leq \epsilon, \nonumber
\end{align}
which limits the Wasserstein distance between the current- and next context distribution $q(\svec{c})$ and $p(\svec{c})$. \rebuttal{Please note that we overload the meaning of the symbol $\epsilon$ with step size for \gradient and the trust region for \currot, as both concepts limit the change in sampling distribution between updates.} We realize the performance estimator using Nadaraya-Watson kernel regression \citep{nadaraya1964estimating,watson1964smooth} with a squared exponential kernel
\begin{align*}
    \hat{J}(\pi, \svec{c}) {=} \frac{\sum_{l=1}^L K_h(\svec{c}, \svec{c}_l) R_l}{\sum_{l=1}^L K_h(\svec{c}, \svec{c}_l)}, \; K_h(\svec{c}, \svec{c}_l) {=} \exp\left(-\frac{d(\svec{c}, \svec{c}_l)^2}{2 h^2}\right).
\end{align*}
This estimator does not rely on gradient-based updates and requires no architectural choices except for the lengthscale $h$, consequently not complicating the application of the overall algorithm. We postpone the discussion of this lengthscale parameter $h$ until after we have discussed the approximate optimization of Objective (\ref{eq:currot:approximate-currot}) and first focus on the choice of dataset $\mathcal{D} {=} \{(\svec{c}_l, R_l) | l \in [1,L]\}$ used to build the kernel regressor. \\
We create the dataset from two buffers, \rebuttal{$\mathcal{D}_+$ and $\mathcal{D}_-$}, of size~$N$. We update the buffers with the results of policy rollouts $(\svec{c}, R_{\svec{c}})$ during agent training, where $R_{\svec{c}} {=} \sum_{t=0}^{\infty} \gamma^t r_{\svec{c}}(\svec{s}_t, \svec{a}_t)$. While \rebuttal{$\mathcal{D}_-$} is simply a circular buffer that keeps the most recent $N$ rollouts with $R_{\svec{c}}$ below the performance threshold $\delta$, \rebuttal{$\mathcal{D}_+$} contains contexts $\svec{c}$ for which $R_{\svec{c}} \geq \delta$. However, \rebuttal{$\mathcal{D}_+$} is updated differently if full. Once full, we interpret the samples in \rebuttal{$\mathcal{D}_+$} as an empirical distribution $\hat{p}_+(\svec{c})$ and select rollouts from the union of \rebuttal{$\mathcal{D}_+$} and the set of new rollouts above the performance threshold $\delta$ to minimize $\mathcal{W}_2(\hat{p}_+, \hat{\mu})$. This optimal selection can be computed with a generalized version of the optimal assignment problem \rebuttal{(\ref{eq:currot:assigment}), where $\hat{p}_+$ is represented by $N_+$ particles and $\hat{\mu}$ is represented by $N$ particles with $N_+ \geq N$. The generalized problem then produces a selection of $N$ particles to represent $\hat{p}_+$, which minimizes the resulting distance $\mathcal{W}(\hat{p}_+, \hat{\mu})$}. We can hence interpret $\hat{p}_+(\svec{c})$ as a conservative solution to the \currot objective (\ref{eq:currot:currot}). The solution is conservative since the particles are obtained from past iterations and may exceed the performance threshold $\delta$ by some margin, hence not targeting the exact border of the performance threshold.\begin{algorithm}[t]
\caption{Approximate \currot}\label{alg:currot:currot}
\begin{algorithmic}
\STATE {\bfseries Input:} Initial context dist.\ $p_0(\svec{c})$, target context dist.\ $\mu(\svec{c})$, metric $d(\svec{c}_1{,} \svec{c}_2)$, performance bound $\delta$, distance bound~$\epsilon$
    \STATE {\textbf{Initialize:} $\hat{p}(\svec{c}) = \frac{1}{N} \sum_{n=1}^N \delta_{\svec{c}_{p_0, n}}(\svec{c}),\ \svec{c}_{p_0, n} \sim p_0(\svec{c})$}
    \WHILE{True}
    \STATE {\bfseries Agent Improvement:}
   \STATE {Sample contexts $\svec{c}_m \sim \hat{p}(\svec{c}),\ m \in [1, M]$}
   \STATE {Train policy $\pi$ under $\svec{c}_m$ and observe episodic rewards $R_m = \sum_{t=0}^\infty \gamma^t r_{\svec{c}_m}(\svec{s}_t, \svec{a}_t),\ m \in [1, M]$}
   \STATE {\bfseries Context Distribution Update:}
   \STATE Update buffers \rebuttal{$\mathcal{D}_+$ and $\mathcal{D}_-$} with $\{ (\svec{c}_m, R_m) \vert m {\in} [1, M] \}$
   \STATE Estimate $\hat{J}(\pi, \svec{c}) \approx J(\pi, \svec{c})$ from \rebuttal{$\mathcal{D}_+$ and $\mathcal{D}_-$}
   \STATE Update $\hat{p}(\svec{c})$ via Eq. (\ref{eq:currot:currot-particle-opt}) and $\hat{J}(\pi, \svec{c})$, $\hat{p}(\svec{c})$, $\hat{\mu}(\svec{c})$
   \ENDWHILE
\end{algorithmic}
\end{algorithm}\\
To more precisely target this border of agent competence, we proceed as follows: First, we solve an assignment problem between $\hat{p}(\svec{c})$ and $\hat{p}_+(\svec{c})$ to obtain pairs $(\svec{c}_{p,n}, \svec{c}_{p_+,\rebuttal{\phi}(n)})$. We then reset $\svec{c}_{p,n} {=} \svec{c}_{p_+,\rebuttal{\phi}(n)}$ for those contexts $\svec{c}_{p,n}$ with $\hat{J}(\pi,\svec{c}_{p,n}) {<} \delta$. Next, we again sample an empirical target distribution $\hat{\mu}(\svec{c})$ and solve an assignment problem between the updated empirical distribution $\hat{p}(\svec{c})$ and $\hat{\mu}(\svec{c})$ to obtain context pairs $(\svec{c}_{p,n}, \svec{c}_{\mu,\rebuttal{\phi}(n)})$. We then solve an optimization problem for each pair to obtain the particles for the new empirical context distribution
\begin{align}
    \argmin_{\svec{c} \in \mathcal{C}}&\ d(\svec{c}, \svec{c}_{\mu,\rebuttal{\phi}(n)}) \label{eq:currot:currot-particle-opt} \\
    \text{s.t.}&\ \hat{J}(\pi, \svec{c}) \geq \delta \qquad d(\svec{c}, \svec{c}_{p,n}) \leq \epsilon. \nonumber
\end{align}
Note that the restriction $d(\svec{c}, \svec{c}_{p,n}) {\leq} \epsilon$ ensures that $\mathcal{W}_2(\hat{p}, \hat{q}) {\leq} \epsilon$, while de-coupling the optimization for the individual particles. \rebuttal{We use a simple approximate optimization scheme that samples a set of candidate contexts around $\svec{c}_{p,n}$ and selects the candidate that minimizes the distance to $\svec{c}_{\mu,\phi(n)}$ while fulfilling the performance constraint. In the continuous Euclidean settings, we uniformly sample candidates in the half ball of contexts that make an angle of less than 90 degrees with the descent direction $\svec{c}_{p,n} - \svec{c}_{\mu, \phi(n)}$. In discrete context spaces, we evaluate all contexts in the trust region. If even after resetting $\svec{c}_{p,n} {=} \svec{c}_{p_+,\rebuttal{\phi}(n)}$, no candidate satisfies the performance threshold, and hence Objective (\ref{eq:currot:currot-particle-opt}) is infeasible, we set $\svec{c}_{p,n}$ to the candidate with maximum performance in the $\epsilon$-ball.} \\
Having defined Objective (\ref{eq:currot:currot-particle-opt}), we can discuss the lengthscale parameter $h$ of the Nadaraya-Watson estimator. Given that the purpose of the estimator is to capture the trend in the $\epsilon$-ball around a particle $\svec{c}_{p,n}$, we simply set the lengthscale to $0.3 \epsilon$. This choice ensures that the two-times standard deviation interval of the squared-exponential kernel $K_h$ centered on $\svec{c}_{p,n}$ covers the trust region. \\
\rebuttal{Like for \gradient, we train on $p_0(\svec{c})$ until reaching an average performance of at least $\delta$, at which point we update the distribution according to Algorithm \ref{alg:currot:currot}.}

\section{Experiments}

To demonstrate the behavior of the introduced algorithms \currot and \gradient, we benchmark the algorithms in different environments that feature discrete- and continuous context spaces with Euclidean- and non-Euclidean distance metrics. We furthermore evaluate both the exact approaches as well as their approximate implementations. To highlight the benefits of the proposed approach over currently popular CRL methods, we compare against a range of baselines. More precisely, we evaluate \alpgmm \citep{portelas2019teacher}, \goalgan \citep{florensa2018automatic}, \plr \citep{jiang2021prioritized}, \vds \citep{zhang2020automatic} and \acl \citep{graves2017automated} in addition to a random curriculum and training directly on $\mu(\svec{c})$ (referred to as Default). Details of the experiments, such as hyperparameters and employed RL algorithms, can be found in Appendix \ref{app:currot:experiments}. The code for running the experiments will be made publicly available upon acceptance.

\subsection{E-Maze Environment}
\label{sec:currot:e-maze}

\begin{figure}[t]
    \centering
    \includegraphics{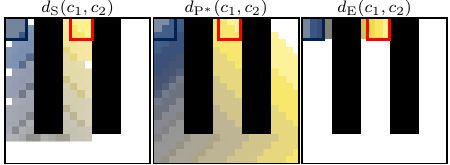}
    \caption{E-Maze environment and visualizations of barycenters between initial- and target task distribution for the shortest-path distance $d_{\text{S}}$, performance pseudo-distance $d_{\text{P}^*}$ and Euclidean distance $d_{\text{E}}$. Brighter colors correspond to distributions generated at later stages of the interpolation. The states covered by initial- and target task distributions are highlighted by the blue and red lines.}
    \label{fig:currot:emaze-metrics}
\end{figure}

\begin{figure}[b!]
    \centering
    \includegraphics{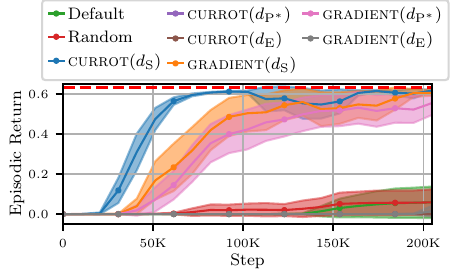}
    \caption{Expected return on the target task distribution $\mu(c)$ in the E-Maze environment achieved by \currot and \gradient under varying distance metrics. The shaded area corresponds to two times the standard error (computed from $20$ seeds). The red dotted line represents the maximum possible reward achievable on $\mu(c)$.}
    \label{fig:currot:emaze-performance}
\end{figure}

\begin{figure}[t]
    \centering
    \includegraphics{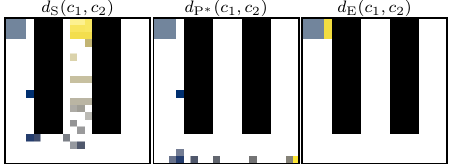}
    \caption{\currot sampling distribution without entropy regularization for varying distance measures. Brighter colors correspond to later training iterations.}
    \label{fig:currot:emaze-currot-interpolations}
\end{figure}

\begin{figure}[b!]
    \centering
    \includegraphics{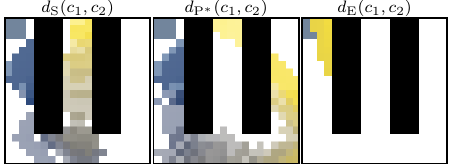}
    \caption{\currot sampling distribution for $H_{\text{LB}} {=} 2$ and varying distance measures. Brighter colors correspond to later training iterations.}
    \label{fig:currot:emaze-entropy-ablation}
\end{figure}

To investigate \currot and \gradient without relying on approximations and highlight the effect of the chosen distance metric, we start the experiments with the environment shown in Figure \ref{fig:currot:emaze-metrics}. \rebuttal{In this sparse-reward environment that is represented by a $20 \times 20$ grid, an agent is tasked to reach a goal position by moving around an elongated wall (black tiles in Figure \ref{fig:currot:emaze-metrics}). The curricula for this task control the goal position to be reached via the context $c$.} We investigate three different distance functions of $\mathcal{C}$ in this environment:
\begin{itemize}
    \item A Euclidean distance $d_{\text{E}}(c_1, c_2) {=} \| \svec{r}(c_1) - \svec{r}(c_2) \|_2$ based on representations $\svec{r}(c) \in \mathbb{R}^3$ of the discrete contexts which encode the two-dimensional goal position as well as the height (walls have a height of $200$ and regular tiles a height of zero).
    \item A shortest-path distance $d_{\text{S}}(c_1, c_2)$ computed using the Dijkstra algorithm. The search graph for the Dijkstra algorithm is built by connecting neighboring contexts using the previously defined Euclidean distance.
    \item A pseudo-metric investigated by \citet{huang2022curriculum} that is based on the optimal policy's absolute difference in expected return $d_{\text{P}^*}(c_1, c_2) {=} \left| J_{\pi^*}(c_1) - J_{\pi^*}(c_2) \right|$. \rebuttal{Opposed to the metrics $d_{\text{E}}$ and $d_{\text{S}}$, this pseudo-metric can assign $d_{\text{P}^*}(c_1, c_2)=0$ for $c_1 \neq c_2$.}
\end{itemize}
\rebuttal{While the definition of Wasserstein barycenters is not entirely rigorous for the pseudo-metric $d_{\text{P}^*}$, the introduced approximate algorithms can still operate on it without problems}. \citet{huang2022curriculum} also investigated this pseudo-metric for the current policy $\pi$, leading to a different metric in each algorithm iteration. We investigate this interesting concept in Appendix \ref{app:currot:experiments-emaze} to keep the main article short and consistent with the previous sections that assumed a fixed distance. Figure \ref{fig:currot:emaze-metrics} visualizes the barycentric interpolations generated by $d_{\text{E}}$, $d_{\text{S}}$, and $d_{\text{P}^*}$. Looking at Figure \ref{fig:currot:emaze-metrics}, we can already anticipate a detrimental effect of the Euclidean metric $d_{\text{E}}$ on the generation of the curriculum. The visualization of $d_{\text{P}^*}$ indicates a weakness of purely performance-based metrics since a similar expected return for $c_1$ and $c_2$ does not guarantee similar outcomes of actions in the two contexts. \\
We visualize the expected return for different curricula in Figure \ref{fig:currot:emaze-performance}.
As we can see, \currot and \gradient can significantly improve performance over both a purely random- as well as no curriculum. However, the performance gains are highly dependent on an appropriate choice of metric. While both \currot and \gradient show strong performance for $d_{\text{S}}$, \currot's performance diminishes for $d_{\text{P}^*}$, and none of the two methods can make the agent proficient on $\mu(c)$ when using $d_{\text{E}}$. \\
Figure \ref{fig:currot:emaze-currot-interpolations} shows interpolations generated by \currot for the investigated metrics. We see that the interpolating distributions of \currot can collapse to a Dirac distribution for $d_{\text{S}}$ and $d_{\text{P}^*}$. As discussed in Section \ref{sec:currot:approx_currot}, \citet{huang2022curriculum} proposed using an entropy-regularized version of optimal transport due to its computational speed. Given that we solve Objectives (\ref{eq:currot:currot}) and (\ref{eq:currot:gradient}) analytically, we can investigate the effect of entropy-regularization not with respect to computational speed but to performance. In Table \ref{table:currot:entropy-ablation}, we show the final agent performance when using entropy-regularized transport plans for \gradient as well as a lower bound $H_{\text{LB}}$ on the entropy of the generated task distributions for \currot.\begin{table}[t!]
\renewcommand{\arraystretch}{1.3}
\caption{Final agent performance of \currot and \gradient on $\mu(c)$ in the E-Maze environment for varying amounts of entropy regularization ($\lambda$ and $H_{\text{LB}}$). Mean and standard error are computed from $20$ seeds.}
\label{table:currot:entropy-ablation}
\centering
\begin{tabular}{c|c|c|c|c}
\multicolumn{5}{c}{\currot} \\
\hline
$H_{\text{LB}}$ & $0.$ & $0.5$ & $1.0$ & $2.0$ \\
\hline
$d_{\text{S}}$ & $0.62 {\pm} 0$ & $0.61 {\pm} 0$ & $0.53 {\pm} 0.04$ & $0.58 {\pm} 0.03$\\
\hline
$d_{\text{P}^*}$ & $0 {\pm} 0$ & $0.45 {\pm} 0.06$ & $0.38 {\pm} 0.06$ & $0.42 {\pm} 0.06$\\
\hline
$d_{\text{E}}$ & $0 {\pm} 0$ & $0 {\pm} 0$ & $0 {\pm} 0$ & $0 {\pm} 0$\\
\hline
\multicolumn{5}{c}{\gradient} \\
\hline
$\lambda$ & $0.$ & $10^{-8}$ & $10^{-4}$ & $10^{-2}$ \\
\hline
$d_{\text{S}}$ & $0.60 {\pm} 0.01$ & $0.56 {\pm} 0.04$ & $0.62 {\pm} 0.00$ & $0.60 {\pm} 0.01$ \\
\hline
$d_{\text{P}^*}$ & $0.55 {\pm} 0.03$ & $0.48 {\pm} 0.05$ & $0.45 {\pm} 0.05$ & $0.30 {\pm} 0.06$ \\
\hline
$d_{\text{E}}$ & $0.01 {\pm} 0.01$ & $0.03 {\pm} 0.03$ & $0.03 {\pm} 0.03$ & $0.01 {\pm} 0.01$
\end{tabular}
\end{table} The detailed formulations of these variants are provided in Appendix \ref{app:currot:experiments-emaze}. As the results show, entropy regularization can benefit \currot. The visualizations in Figure \ref{fig:currot:emaze-entropy-ablation} indicate that this benefit arises from avoiding the aggressive targeting of contexts right at the edge of the performance constraint that we can see in Figures \ref{fig:currot:pull_figure}, \ref{fig:currot:currot_gradient_interpolations}, and \ref{fig:currot:emaze-currot-interpolations}. In the case of the pseudo distance $d_{\text{P}^*}$, the more diverse tasks sampled from $p(c)$ sometimes allowed the agent to generalize enough to solve tasks sampled from $\mu(c)$. For \gradient, we cannot see significant performance gains but can observe that a too-high entropy regularization in combination with $d_{\text{P}^*}$ diminished performance. Given that for an adequate metric (i.e., $d_{\text{S}}$), the observed performance is stable across different amounts of entropy regularization, we do not further explore this avenue in the following experiments.

\subsection{Unlock-Pickup Environment}
\label{sec:currot:unlockpickup}

\begin{figure}[b!]
    \centering
    \includegraphics{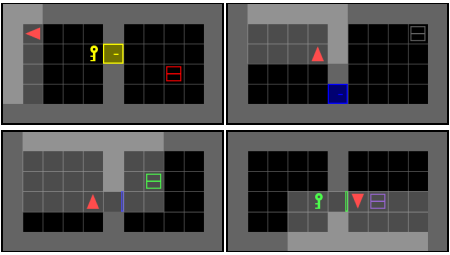}
    \caption{The Unlock-Pickup environment, in which an agent needs to pick up the box in the right room by unlocking the door. After reset, the agent is randomly placed in the left room not carrying the key (top left image). After picking up the key (top right), the door can be unlocked (bottom left) to move to the box (bottom right). The door-, box- and key positions as well as their colors vary across environment resets. The agent receives a partial view of the world (highlighted rectangle) that is blocked by walls and closed doors.}
    \label{fig:currot:unlockpickup-env}
\end{figure}

In the following environment, we aim to benchmark approximate implementations of \currot and \gradient for large discrete context spaces and demonstrate that appropriate distances for non-trivial context spaces can be designed by hand. In Figure \ref{fig:currot:unlockpickup-env}, we visualize the unlock-pickup environment from the Minigrid environment collection \citep{minigrid} that we chose for this investigation. To master this environment, the agent must pick up a key, unlock a door and eventually pick up a box in the room that has just been unlocked. \\
We define a curriculum by controlling the starting state of an episode via the context $c$, i.e., controlling the position of the box, key, agent, and door, as well as the state of the door (whether closed or open). As detailed in Appendix \ref{app:currot:experiments-unlockpickup}, this task parameterization results in $81.920$ tasks to compile a curriculum from. The initial context distribution is defined to encode states in which the agent is directly in front of the box, similar to the bottom-right image in Figure \ref{fig:currot:unlockpickup-env}. Starting from this initial distribution, the learning algorithm needs to generate a curriculum that ultimately allows the agent to reach and pick up the box from a random position in the left room with a closed door. As we show in Appendix \ref{app:currot:experiments-unlockpickup}, it is possible to define a so-called highway distance function \citep{baikousi2011similarity} between contexts that properly takes the role of the door and its interaction with the key into account, without relying on a planning algorithm like in the previous environment. We use this distance function in the following evaluations.%
\begin{figure}[t]
    \centering
    \includegraphics{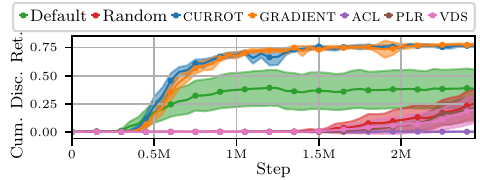}
    \caption{Episodic return on the target task distribution $\mu(c)$ in the Unlock-Pickup environment for different curricula. The shaded area corresponds to two times the standard error computed from $20$ seeds.}
    \label{fig:currot:unlockpickup-performance}
\end{figure}\begin{figure*}[b!]
    \centering
    \begin{subfigure}[t]{0.195 \textwidth}
    \centering
    \includegraphics{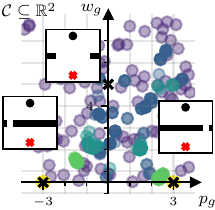}
    \caption{\currot}
	\end{subfigure}
    \begin{subfigure}[t]{0.17 \textwidth}
    \centering
    \includegraphics{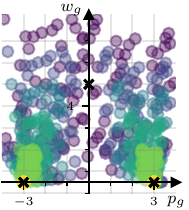}
    \caption{\gradient}
	\end{subfigure}
    \begin{subfigure}[t]{0.59 \textwidth}
    \centering
    \includegraphics{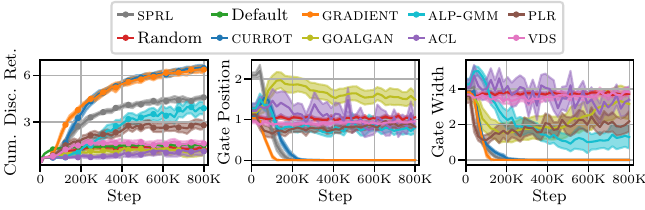}
    \caption{Performance Comparison}
	\end{subfigure}
    \caption{a + b) The point-mass environment with its two-dimensional context space. The target distribution $\mu(\svec{c})$ encodes the two gates with width $w_g {=} 0.5$, in which the agent (black dot) is required to navigate through a narrow gate at different positions to reach the goal (red cross). The colored dots visualize a curriculum generated by \currot and \gradient for this environment. c) Left: Discounted cumulative return over learning epochs obtained in the point mass environment under different curricula as well as baselines that sample tasks uniformly from all of $\mathcal{C}$ (Random) or $\mu(\svec{c})$ (Default). Middle and Right: Median minimum distance to the target contexts of $\mu(\svec{c})$ for the two dimensions of the context space (i.e., gate position and -width). Mean and two-times standard error intervals are computed from $20$ seeds.}
    \label{fig:currot:point-mass-performance}
\end{figure*}\\
In addition to the approximate versions of \currot and \gradient, we evaluate \plr, \vds, and \acl on this task. We do not evaluate \sprl, \alpgmm, and \goalgan since those algorithms have been designed for continuous and Euclidean context spaces by, e.g., leveraging Gaussian distributions, kd-trees, or Gaussian sampling noise. The evaluation results in Figure \ref{fig:currot:unlockpickup-performance} show that \currot and \gradient consistently allow mastering the target tasks (a cumulative discounted return of $0.75 \approx 0.99^{28}$ is obtained by solving a task in $28$ steps). For both \currot and \gradient, each of the $20$ runs led to a well-performing policy, and we can barely see any difference in learning speed between the approaches. Learning directly on the target task distribution allows mastering the environment in some runs while failing to do so in others due to the high dependence on collecting enough positive reward signals at the beginning of learning. These two outcomes lead, on average, to a lower performance compared to \currot and \gradient. Finally, we see that all baseline curriculum methods learn slower than directly learning on the target task distribution $\mu(c)$, with \acl not producing policies that collect any reward on the target tasks. Given the successful application of \plr in the Procgen benchmark, which features a diverse set of Arcade game levels with highly distinct visual observations, we wish to discuss the observed low performance of \plr here in more detail. As we show in Appendix \ref{app:currot:experiments-unlockpickup}, \plr indeed samples contexts occurring under $\mu(c)$ with at least $7\%$ in each run. Furthermore, in about half of the runs, the agent also learns to solve those target tasks that are replayed by \plr at some point in the curriculum. However, these replayed target tasks only make up a small fraction of the total number of target tasks, resulting in low performance on all of $\mu(c)$. The absence of a notion of target distribution for \plr seems to lead to ineffective use of samples w.r.t improving performance on the target. This lack of target distribution causing problems will be a re-occurring theme for the subsequent experiments.

\subsection{Point-Mass Environment}
\label{sec:currot:point-mass}

\begin{figure*}[t]
    \centering
    \begin{subfigure}[t]{0.17 \textwidth}
    \centering
    \includegraphics{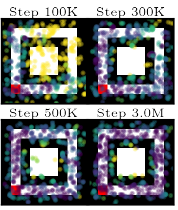}
    \caption{\currot }
	\end{subfigure}
    \begin{subfigure}[t]{0.2 \textwidth}
    \centering
    \includegraphics{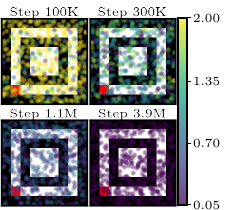}
    \caption{\gradient}
	\end{subfigure}
    \begin{subfigure}[t]{0.62 \textwidth}
    \centering
    \includegraphics{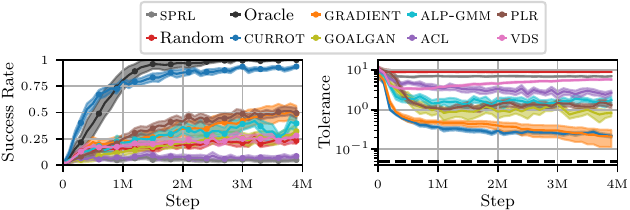}
    \caption{Performance Comparison}
	\end{subfigure}
    \caption{a + b) Curricula generated by \currot and \gradient in the spare goal-reaching (SGR) environment at different epochs. The starting area of the agent is highlighted in red. The walls are shown in black. The position of the samples encodes the goal to be reached while the color encodes the goal tolerance. c) Success rate on the feasible subspace of $\mathcal{C}$ (left) and median goal tolerance (right) for different CRL methods in the SGR environment. We also include an oracle baseline that only samples the feasible tasks in the context space $\mathcal{C}$. For both plots, mean and two-times standard error intervals are computed from $20$ runs.}
    \label{fig:currot:o-maze-performance}
\end{figure*}

In this environment, in which a point-mass agent must pass through a narrow gate to reach a goal position opposite a wall (Figure \ref{fig:currot:point-mass-performance}), we benchmark our approximate implementations of \currot and \gradient in continuous settings. The context $\svec{c} {\in} \mathbb{R}^2$ controls the position and width of the gate that the agent needs to pass. This environment has been introduced with the \sprl algorithm by \citet{klink2019self} with a Gaussian target distribution that essentially encodes one narrow gate requiring the agent to detour before reaching the target position. Combined with a dense reward based on the Euclidean distance to the goal, the target task is subject to a prominent local minimum that simply moves the agent close to the wall without passing through. We extend this task with a bi-modal target distribution that challenges \sprl's Gaussian restriction that -- as we discussed -- is required for it to work properly. As seen in Figure \ref{fig:currot:point-mass-performance}, \currot and \gradient generate curricula that target both modes of the distribution and allow learning a proficient policy on all of $\mu(\svec{c})$. 
As we show in Appendix \ref{app:currot:experiments-point-mass}, the Gaussian restriction of \sprl's context distribution leads to $p(\svec{c})$ matching only one of the modes of $\mu(\svec{c})$, resulting in a lower average reward on $\mu(\svec{c})$ compared to \currot and \gradient. We additionally visualize summary statistics for the other CRL methods in Figure \ref{fig:currot:point-mass-performance}, showing that they result in a less targeted sampling of contexts likely under $\mu(\svec{c})$. This observation, in combination with the lower performance compared to \currot and \gradient, once more emphasizes the importance of embedding a notion of target distribution in CRL algorithms. \\
\rebuttal{We additionally benchmark \currot and \gradient in versions of the point-mass environment with increasing context spaces dimensions. The results in Appendix \ref{app:currot:high-dim} show that both approaches can scale to higher dimensions (we investigated up to $30$-dimensional context spaces) for this environment. However, they also emphasize the importance of certain algorithmic choices such as the choice of initial context distribution $p_0(\svec{c})$ for both methods and the choice of the trust region as well as the sampling schemes of candidates for Objective (\ref{eq:currot:currot-particle-opt}) for \currot. To keep the main article short, we refer the interested reader to Appendix \ref{app:currot:high-dim}.}

\subsection{Sparse Goal-Reaching Environment}

We next turn to a sparse-reward, goal-reaching environment in which an agent needs to reach a desired position with high precision (Figure \ref{fig:currot:o-maze-performance}). Such environments have, e.g., been investigated by \citet{florensa2018automatic}. The context $\svec{c} {\in} \mathcal{C} \subseteq \mathbb{R}^3$ of this environment encodes the 2D goal position as well as the allowed tolerance for reaching the goal. This parameterization results in both infeasible tasks being part of $\mathcal{C}$ (unreachable regions) as well as tasks that are solely meant to be stepping stones to more complicated ones (low-precision tasks). Given that the agent is ultimately tasked to reach as many goals as possible with the highest precision, i.e., the lowest tolerance, the target distribution $\mu(\svec{c})$ is a uniform distribution on a 2D slice of $\mathcal{C}$ \revision{with minimal task tolerance}{in which the tolerance of each context is minimal}. The walls in the environment (Figure \ref{fig:currot:o-maze-performance}) render many \revision{target tasks}{tasks encoded by $\mu(\svec{c})$} infeasible, requiring the curriculum to identify the feasible subspace of tasks to achieve a good learning performance. Figure \ref{fig:currot:o-maze-performance} shows that \currot results in the best learning performance across all evaluated CRL methods. Only an oracle, which trains the learning agent only on the feasible subspace of high-precision tasks, can reach higher \revision{performance}{precision}. The evolution of the task tolerances shown in Figure \ref{fig:currot:o-maze-performance} highlights that \currot and \gradient continuously \revision{reduce the task tolerance}{increase the precision with which the goals must be reached}. The baseline CRL methods lack focus on the tasks encoded by $\mu(c)$, sampling tasks with comparatively high tolerance even towards the end of training.
Interestingly, \sprl \revision{samples high-tolerance tasks throughout all training epochs since its Gaussian context distribution converges}{does not progress to high-precision tasks but continues to sample tasks of high tolerance in later training epochs. As we show in Appendix \ref{app:currot:experiments-sgr}, this behavior is caused by the Gaussian context distribution of \sprl converging} to a quasi-uniform distribution over $\mathcal{C}$. Otherwise, \sprl would not be able to cover the non-Gaussian \revision{}{target} distribution of feasible high-precision \revision{target}{} tasks without encoding many infeasible tasks. Figure \ref{fig:currot:o-maze-performance} shows the \revision{particle evolution}{evolution of particles} for runs of \currot and \gradient. \currot gradually decreases the goal tolerance over epochs, starting from contexts that are close to the initial position of the agent. Interestingly, it retains higher tolerance contexts located in the walls of the environment even in later epochs due to the trade-off between sampling high-precision tasks and covering all goal positions. \revision{The pre-determined interpolation of \gradient cannot adjust to infeasible parts of the context space and reduces to a curriculum that shrinks the upper-bound $t_{\text{ub}}$ of the tolerance interval $[0.05, t_{\text{ub}}]$}{The pre-determined interpolation of \gradient cannot adjust to the infeasible parts of the context space and, with that, reduces to a curriculum that shrinks the tolerance interval $[0.05, t_{\text{ub}}]$ by reducing the tolerance $t_{\text{ub}}$}. Consequently, a \revision{decrease in $t_{\text{ub}}$ increases the}{increase in precision goes hand-in-hand with an increasing} number of infeasible tasks on which the agent is trained, slowing down learning and resulting in a significant performance gap between \currot and \gradient in this environment. \revision{We additionally evaluate \currot and \gradient with Hindsight Experience Replay (\her) \cite{andrychowicz2017hindsight} in Appendix \ref{app:currot:experiments-sgr}, showing that \her can serve as a drop-in replacement for \sac in this task.}{}

\vspace{-5pt}
\subsection{Teach My Agent}
\label{sec:currot:tma}

\begin{figure}[t]
\vspace{-3pt}
    \centering
    \begin{subfigure}[b]{0.98 \columnwidth}
		\centering
		\includegraphics{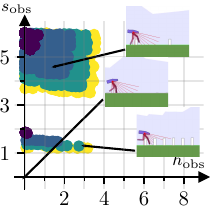}
        \hspace{10pt}
        \includegraphics{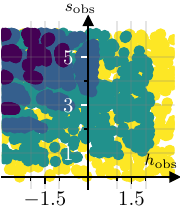}
		\caption{\gradient Curriculum}
	\end{subfigure}
    \begin{subfigure}[b]{0.98 \columnwidth}
		\centering
        \includegraphics{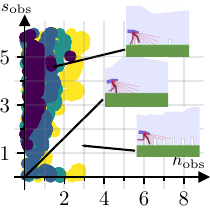}
        \hspace{10pt}
        \includegraphics{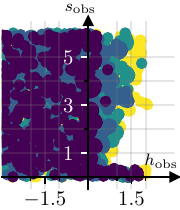}
		\caption{\currot Curriculum}
	\end{subfigure}
    \caption{Sampling distribution of \gradient and \currot on the \textit{teach my agent} benchmark in the \textit{no expert knowledge} setting in task spaces with \textit{mostly infeasible-} (left) and \textit{mostly trivial} (right) tasks. The small images visualize the obstacles encoded by the corresponding contexts. For environment details, please see \citep{romac2021teachmyagent}. Brighter colors indicate tasks at later epochs of training. The yellow dots represent the samples from the last generated distribution.}
    \label{fig:currot:tma_interpolations}
    \vspace{-5pt}
\end{figure}

In this final evaluation environment, a bipedal agent must learn to maneuver over a track of evenly spaced obstacles of a specified height (see Figure \ref{fig:currot:tma_interpolations}). The environment is a modified bipedal walker environment introduced by \citet{portelas2019teacher} and extended by \citet{romac2021teachmyagent} in which the spacing and height of obstacles is controlled by the context $\mathbf{c} \in \mathbb{R}^2$. The evaluations by \citet{romac2021teachmyagent} demonstrated poor performance of \sprl, often performing statistically significantly worse than a random curriculum. Given that both \currot and \gradient can be seen as improved versions of \sprl that -- among other improvements -- explicitly take the geometry of the context space into account, we are interested in whether they can improve upon \sprl. \\
We hence revisit two learning scenarios investigated by \citet{romac2021teachmyagent}, in which CRL methods demonstrated a substantial benefit over random sampling: a setting in which most tasks of the context space are infeasible due to large obstacles and a setting in which most tasks of the context space are trivially solvable. %
\begin{figure*}[t]
    \centering
    \includegraphics{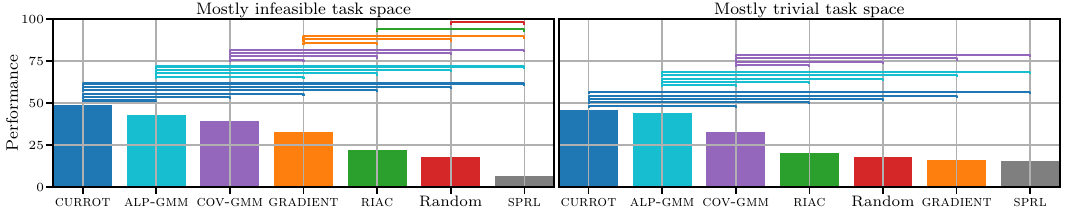}
    \caption{Performance (in percentage of solved tasks) in the \textit{Teach My Agent} benchmark in the \textit{no expert knowledge} setting. The baseline results are taken from \citep{romac2021teachmyagent}, and only \currot and \gradient are evaluated by us. Statistics have been computed from $32$ seeds. Horizontal lines between connecting two methods indicate statistically significant different performances according to Welch's t-test with $p < 0.05$.}
    \label{fig:currot:tma_performance}
\end{figure*}%
Both scenarios lead to slow learning progress when choosing tasks randomly due to frequently encountering too complex or too simple learning tasks. \rebuttal{Given that the uniform initial- and target distribution over the context space lead to poor learning performance, we extend the \currot and \gradient method with a simple randomized search to find areas of $\mathcal{C}$ where the agent achieves returns above $\delta$, similar in spirit to \sprl. We describe this method in Appendix \ref{app:currot:feasible-context-search}.} \\
Figure \ref{fig:currot:tma_performance} visualizes the performance of \currot and \gradient in comparison to other CRL methods that were already evaluated by \citet{romac2021teachmyagent}. We see that \currot achieves the best performance in all environments, in one case performing statistically significantly better than \alpgmm, the best method evaluated in \citep{romac2021teachmyagent}. We also see that the extended version of \gradient can improve upon a random curriculum in the ``mostly infeasible'' scenario while performing insignificantly worse than a random curriculum in the ``mostly trivial'' scenario. Figure \ref{fig:currot:tma_interpolations} can help shed some light on the observed performance difference between \currot and \gradient. For the ``mostly trivial'' scenario, \gradient consistently arrives at sampling from the uniform $\mu(\svec{c})$, whereas \currot focuses on the contexts at the border of agent competence. For the ``mostly infeasible'' scenario, the pre-determined interpolation of \gradient can fail to encode feasible learning tasks, ultimately leading to a lower overall performance than \currot. \\
Summarizing, the experimental results underline that empirically successful curricula can be generated by framing CRL as an interpolation between context distributions. The leap in performance between \gradient and \currot compared to \sprl and the performance differences between \gradient and \currot underline the tremendous impact of design choices, such as the distributional measure of similarity and the way of incorporating performance constraints, on the final algorithm performance. However, when chosen correctly, these curricula exhibit strong performance and allow for guiding training towards tasks of interest specified via $\mu(\svec{c})$. Especially this last aspect can allow for more flexibility in the curriculum design, as it is possible to define auxiliary task parameterizations without jeopardizing learning progress toward tasks of interest. We saw an example of this trade-off in the sparse goal-reaching environment, where the additional precision parameter boosted the performance of \currot while diminishing the performance of other CRL methods.

\section{Conclusion}
In this article, we framed curriculum reinforcement learning as an interpolation between distributions of initial- and target tasks. We demonstrated that the lack of an explicit notion of task similarity in combination with an expected performance constraint makes existing approaches highly dependent on the parameterization of the interpolating task distribution. We avoided these pitfalls by explicitly encoding task similarity via an optimal transport formulation, and by restricting the generated task distributions to only encode tasks that satisfy a specified performance threshold. The resulting method called \currot led to good performance in experiments due to its focus on tasks at the performance threshold and the adaptive nature of the curriculum. Contrasting our approach to a recently proposed method that generates curricula via Wasserstein barycenters between initial- and target task distributions \cite{huang2022curriculum}, we saw that the more adaptive nature of our formulation resulted in better performance when facing learning settings with infeasible target tasks. \rebuttal{In tasks, in which infeasibility is not a concern, both methods performed similar. In Appendix \ref{app:currot:high-dim}, we saw that both methods can scale to higher dimensional tasks although the conceptually more simple \gradient algorithm requires less adaptations of its approximations to do so. Together, both methods demonstrate the benefit of using optimal transport for curriculum RL and we believe that this benefit can be maximized by developing algorithms that combine the adaptivity of \currot with the simpler algorithmic realization of \gradient.} Additionally, we believe that the precise notion of task similarity via the distance $d(\svec{c}_1, \svec{c}_2)$ can prove beneficial in advancing the understanding of curriculum RL. We already saw that an appropriate definition of task similarity is key to successful curriculum learning. We believe that distances learned from experience, which encode a form of intrinsic motivation, will significantly advance these methods by merging the strong empirical results of intrinsic motivation in open-ended learning scenarios \citep{wang2019poet} with the targeted learning achieved by \currot and \gradient.

\ifCLASSOPTIONcompsoc
  \section*{Acknowledgments}
\else
  \section*{Acknowledgment}
\fi

This project has received funding from the DFG project
PA3179/1-1 (ROBOLEAP) and the German Federal Ministry of Education and Research (BMBF) (Project: 01IS22078). This work was also funded by Hessian.ai through the project 'The Third Wave of Artificial Intelligence – 3AI' by the Ministry for Science and Arts of the state of Hessen. Finally, the authors gratefully acknowledge the computing time provided to them on the high-performance computer Lichtenberg at the NHR Centers NHR4CES at TU Darmstadt. This is funded by the Federal Ministry of Education and Research and the state governments participating on the basis of the resolutions of the GWK for national high-performance computing at universities. Joni Pajarinen was supported by Research Council of Finland (formerly Academy of Finland) (decision 345521).

\ifCLASSOPTIONcaptionsoff
  \newpage
\fi

\bibliographystyle{IEEEtranN}
\bibliography{lit.bib}

\begin{IEEEbiography}[{\includegraphics[width=1in,height=1.25in,clip,keepaspectratio]{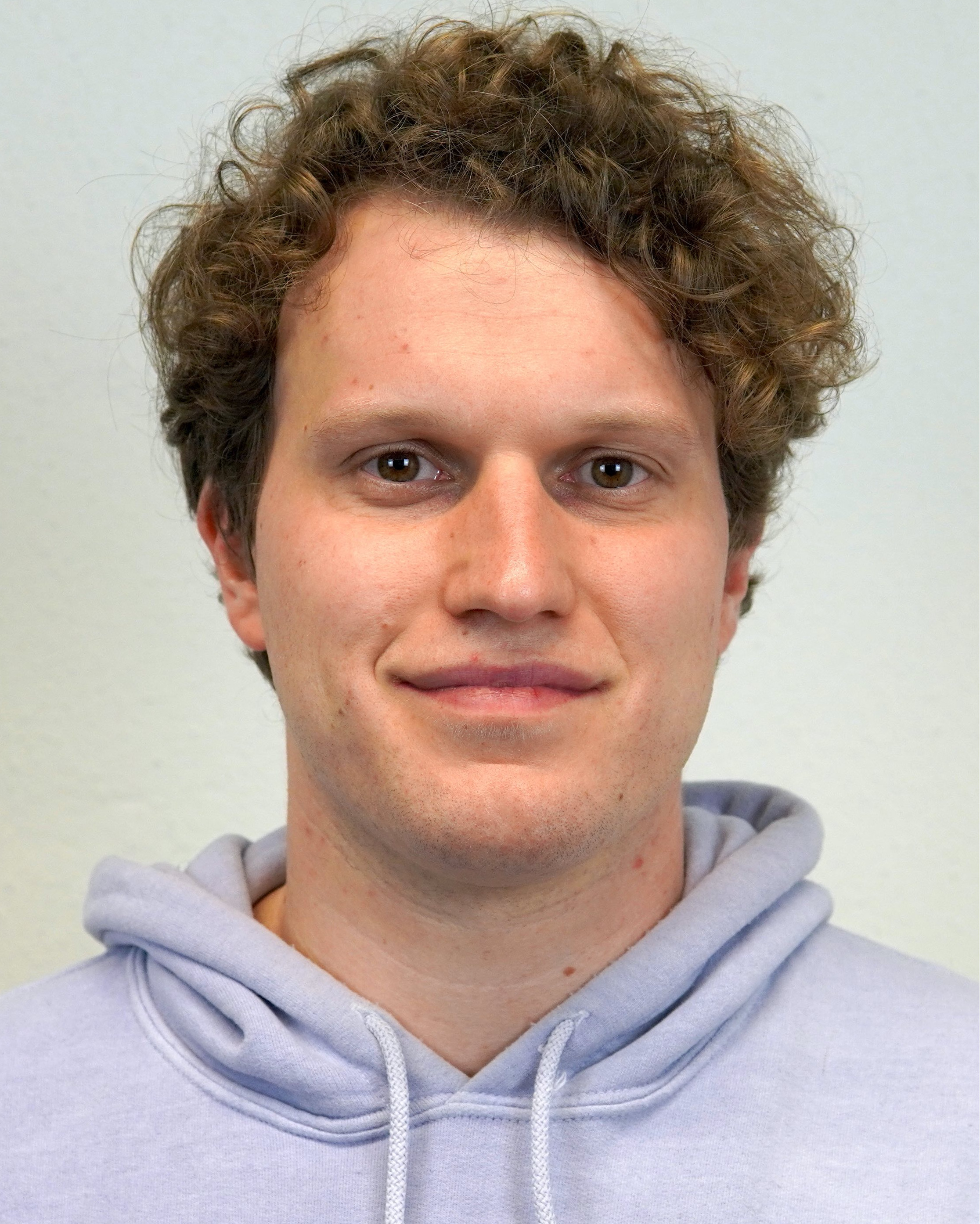}}]{Pascal Klink}
is a Ph.D. student with Jan Peters and Joni Pajarinen at the Institute for Intelligent Autonomous Systems (IAS) at the Techincal University of Darmstadt since May 2019. In his Ph.D., Pascal focuses on improving the learning performance of reinforcement learning agents by leveraging experience across learning tasks via curricula. He completed a research internship at Amazon Robotics and received the AI newcomer award (2021) from the German computer science foundation. Before this, Pascal received his M.Sc. from the Technical University of Darmstadt, where he also worked as a student assistant.
\vspace{-40pt}
\end{IEEEbiography}

\begin{IEEEbiography}[{\includegraphics[width=1in,height=1.25in,clip,keepaspectratio]{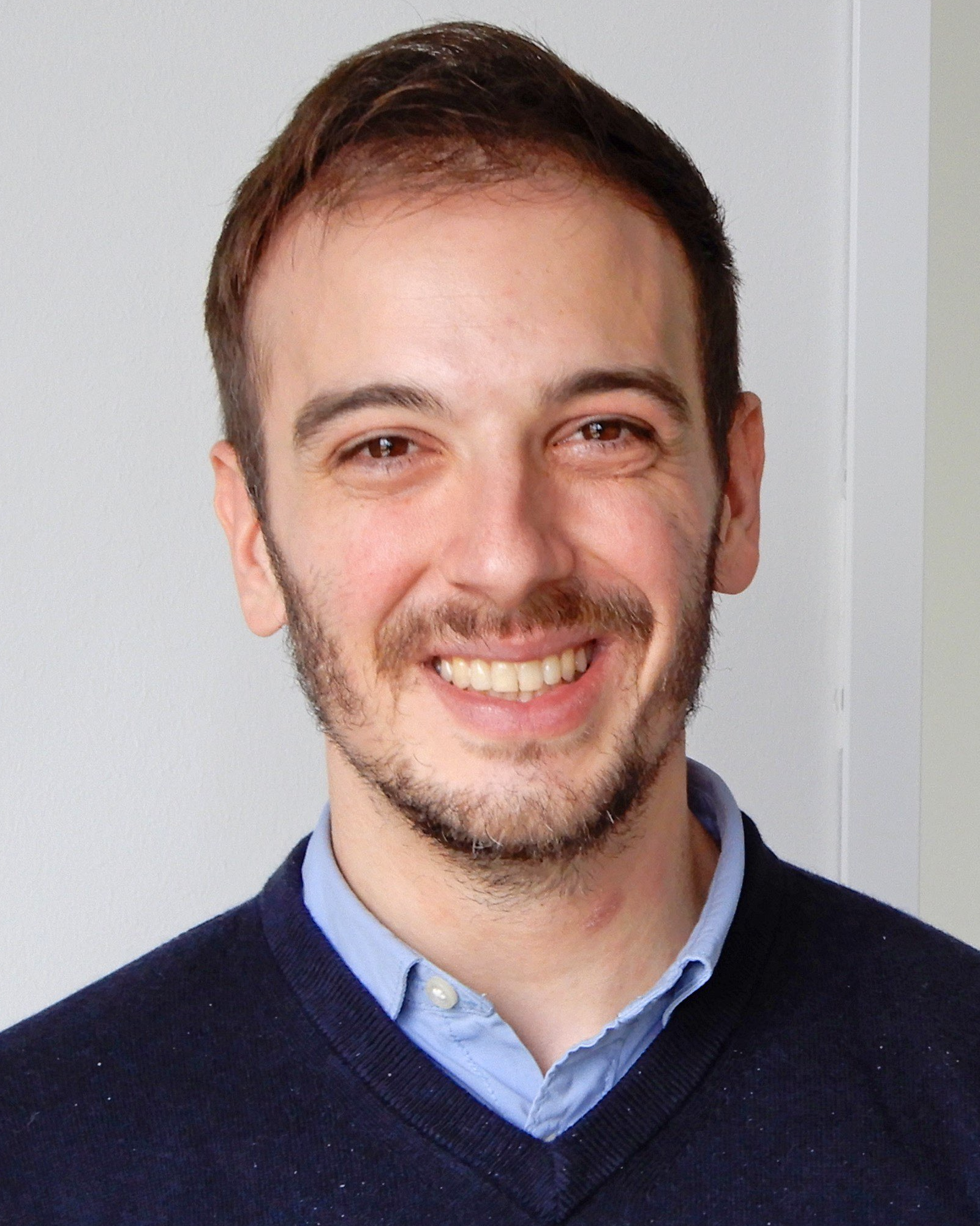}}]{Carlo D'Eramo}
is an Associate Professor for Reinforcement Learning and Computational Decision-Making at the Center for Artificial Intelligence and Data Science of Julius-Maximilians-Universität Würzburg. He is also an independent group leader of the Technical University of Darmstadt and hessian.AI. The research of his LiteRL group revolves around the problem of how agents can efficiently acquire expert skills that account for the complexity of the real world. To answer this question, his group investigates lightweight methods to obtain adaptive autonomous agents, focusing on several RL topics, including multi-task, curriculum, adversarial, options, and multi-agent RL. 
\vspace{-40pt}
\end{IEEEbiography}

\begin{IEEEbiography}[{\includegraphics[width=1in,height=1.25in,clip,keepaspectratio]{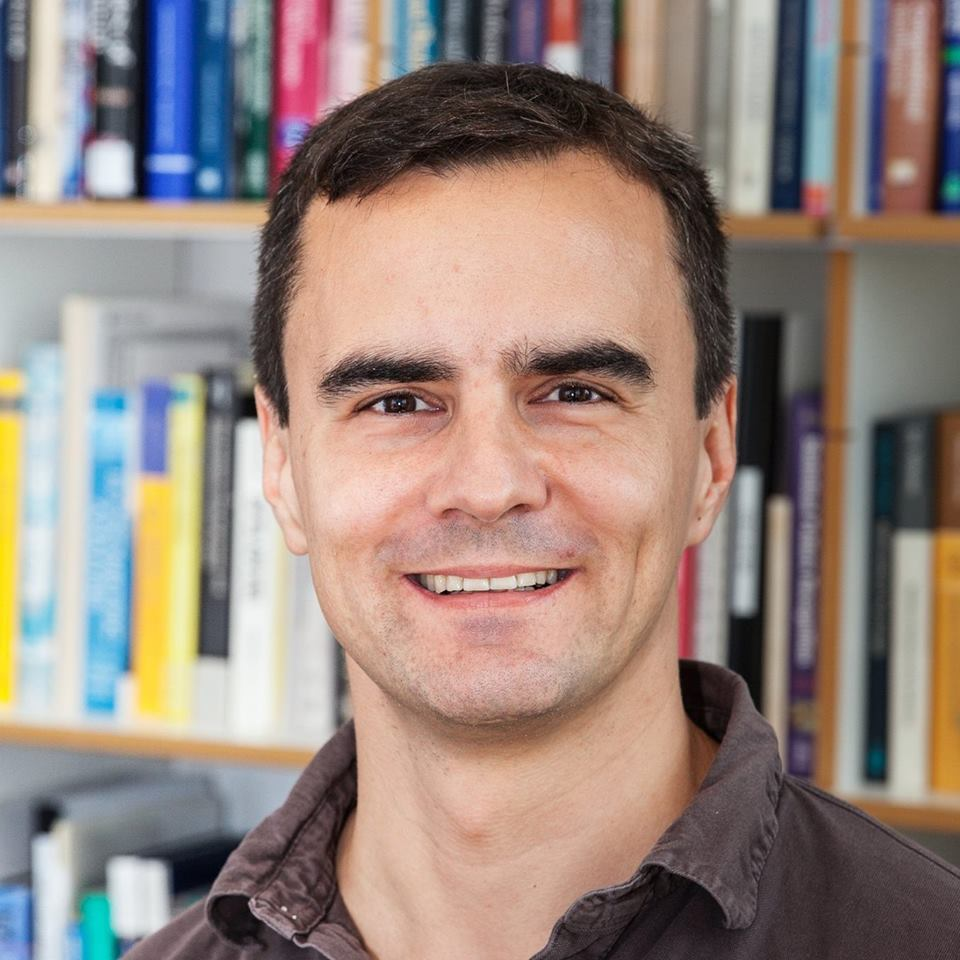}}]{Jan Peters} is a full professor (W3) for Intelligent Autonomous Systems at the Computer Science Department of the Technical University of Darmstadt since 2011 and, at the same time, he is the dept head of the research department on Systems AI for Robot Learning (SAIROL) at the German Research Center for Artificial Intelligence (Deutsches Forschungszentrum für Künstliche Intelligenz, DFKI) since 2022. He is also a founding research faculty member of The Hessian Center for Artificial Intelligence. He has received the Dick Volz Best 2007 US Ph.D. Thesis Runner-Up Award, Robotics: Science \& Systems - Early Career Spotlight, INNS Young Investigator Award, and IEEE Robotics \& Automation Society’s Early Career Award, as well as numerous best paper awards. He received an ERC Starting Grant and was appointed an IEEE fellow, AIAA fellow and ELLIS fellow.
\vspace{-40pt}
\end{IEEEbiography}

\begin{IEEEbiography}[{\includegraphics[width=1in,height=1.25in,clip,keepaspectratio]{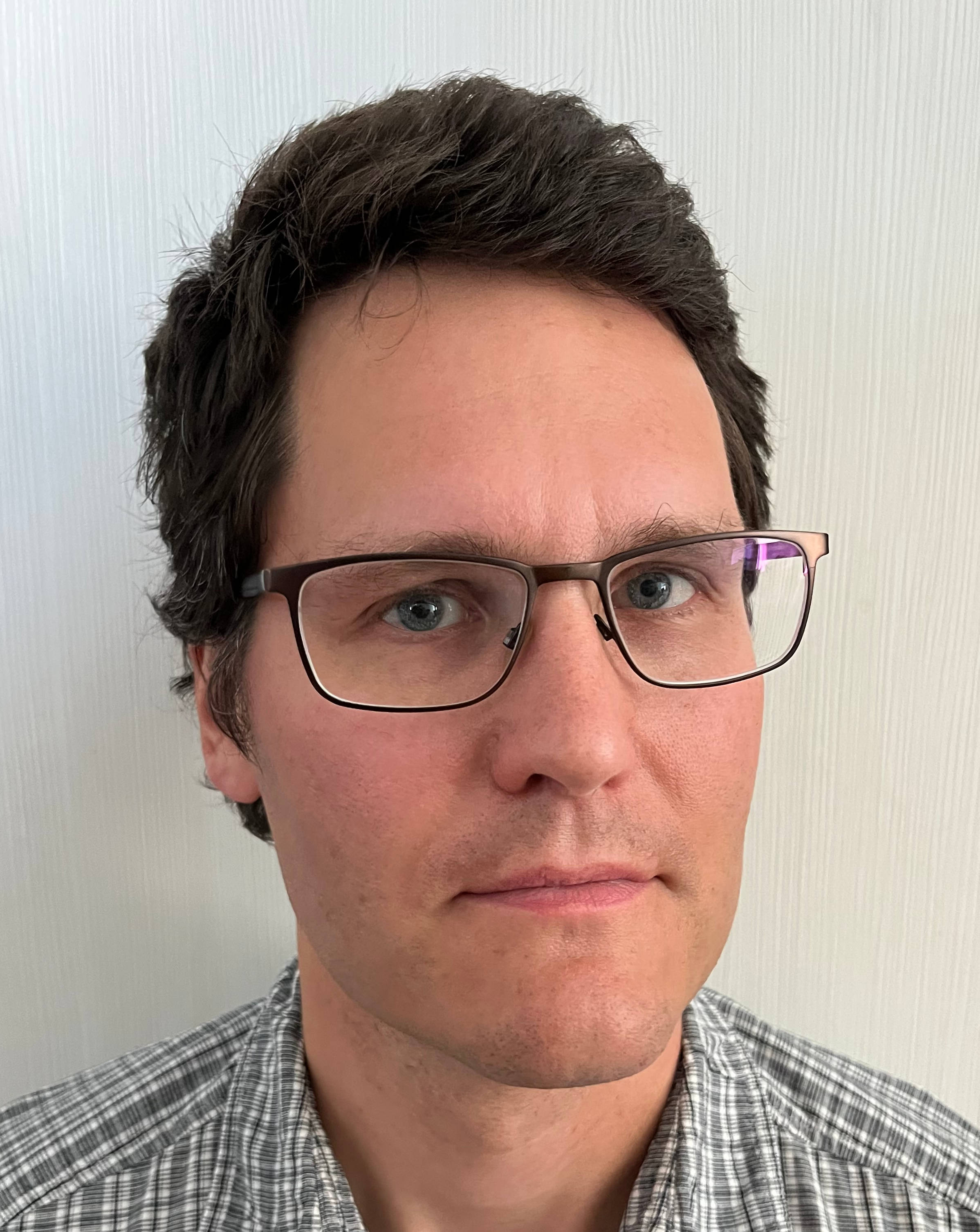}}]{Joni Pajarinen}
is an Assistant Professor at Aalto University, where he leads the Aalto Robot Learning research group. The research group focuses on making robots capable of operating autonomously alongside humans by helping them understand what they need to learn in order to perform their assigned tasks. To this end, the group focuses on developing novel decision-making methods in reinforcement learning, planning under uncertainty, and decision-making in multi-agent systems.
\end{IEEEbiography}

\clearpage
\appendices
\section{Computational Cost of Optimal Transport}
\label{app:currot:ot-complexity}

The benefits of optimal transport (OT), such as explicitly incorporating a ground distance on the sample space, come at the price of a relatively high computational burden caused by the need to solve an optimization problem to compute the Wasserstein distance between two distributions. In practice, OT problems in continuous spaces (such as some of the context spaces investigated in this article) are often reduced to linear assignment problems between sets of particles. Such assignment problems can be exactly solved with variations of the Hungarian algorithm with a time complexity of $\mathcal{O}(n^3)$ \cite{jonker1987shortest}. While this polynomial complexity ultimately leads to prohibitive runtimes for large $n$, we can typically avoid this problem for curriculum RL. Given the often moderate dimensionality of the chosen context spaces, a few hundred particles are typically sufficient to represent the context distributions. In our experiments, we used less than $500$ particles in the continuous environments and $640$ particles for the discrete unlock-pickup environment, leading to observed solving times of less than $200$ms with the \texttt{linear\_sum\_assignment} function of the SciPy library \cite{2020SciPy-NMeth} on an AMD Ryzen 9 3900X. Since the \currot and \gradient algorithms solve, at most, three OT problems per context distribution update, the computational costs of OT are relatively small for the investigated environments. \\
Furthermore, approximations have emerged to tackle problems that require a large number of particles. For example, the GeomLoss library \cite{geomloss}, which we use in the \gradient implementations for continuous Euclidean spaces, implements a variant of entropy-regularized OT that has brought down the computation time of OT for sets of hundreds of thousands of samples to seconds on high-end GPUs \cite{feydy2019interpolating}. So-called sliced Wasserstein distances \cite{bonneel2015sliced,kolouri2019generalized} approximately solve the given OT problem by solving $M$ OT problems in 1-D subspaces, reducing the time complexity to $\mathcal{O}(M n \log(n))$, where typically $M \ll n$. \rebuttal{Finally, neural function approximators have been employed e.g. to speed up the computation of Wasserstein distances by learning a metric embedding from data \cite{courty2018learning} or enable to computation of regularized free-support Wasserstein barycenters by approximating the dual potentials \cite{li2020continuous}.} Consequently, we see opportunities to significantly increase the number of particles via \rebuttal{such approximate} approaches, even though our experiments did not indicate a need for that so far.

\section{Search for Feasible Contexts}
\label{app:currot:feasible-context-search}

As detailed in \rebuttal{Section \ref{sec:currot:tma}}, the initial context distribution $p_0(\svec{c})$ may be uninformed and consequently lead to sampling many learning tasks for which the agent performance is below $\delta$. In such scenarios, we can initiate a search procedure for tasks in which the current agent achieves a performance at least $\delta$ of as long as $\bar{R} = \frac{1}{M}
\sum_{m=1}^M R_m < \delta$. We terminate this search procedure as soon as $\bar{R} \geq \delta$. During this search, \rebuttal{$\mathcal{D}_+$} contains the best-encountered samples, and \rebuttal{$\mathcal{D}_-$} is empty. When a batch of $M$ new episodes arrives, we add those episodes whose return is at least as large as the median return in \rebuttal{$\mathcal{D}_+$} to the buffer -- and for each new episode added, remove the worst performing episode. The search distribution is a (truncated) Gaussian mixture model 
\begin{align*}
    p_{\text{search}}(\svec{c}) &= \sum_{i=1}^{\rebuttal{N_{\mathcal{D}}}} w_i \mathcal{N}\left(\svec{c} \middle| \svec{c}_i, \sigma^ 2_i \mathbf{I}\right)
\end{align*}
with weights $w_i$ and variances $\sigma^2_i$ defined via the minimum return observed over all episodes $R_{\text{min}}$ and the median performance of the buffered episodes $R_{\text{med}}$ 
\begin{align*}
    w_i &\propto \max(0, R_{\svec{c}_i} - R_{\text{med}}), \quad \sigma_i = \max\left(10^{-3}, 2 \frac{\delta - R_{\svec{c}_i}}{\delta - R_{\text{min}}} \right).
\end{align*}
For simplicity of exposition, we assume that $\mathcal{C} = [0,1]^d$, i.e., that the context space is a $d$-dimensional hyper-cube of edge-length one. Consequently, a context $\svec{c}$ with a return of $R_{\text{min}}$ will have a standard deviation of two in each dimension, which, in combination with the Gaussian being truncated, leads to spread-out sampling across the hyper-cube. If the dimensions of $\mathcal{C}$ are scaled differently, a simple re-scaling is sufficient to use the above sampling procedure. \rebuttal{As detailed in the main article, we only required the search procedure in the teach my agent environments, as in the other environments $p_0(\svec{c})$ provided enough successful initial episodes.} For discrete context spaces, the search distribution would need to be adapted, e.g., by defining a uniform distribution over all contexts $\svec{c}$ with a distance $d(\svec{c}, \svec{c}_i)$ less that or equal to a threshold that is similarly scaled as the variance $\sigma^2_i$.

\section{Experimental Details}
\label{app:currot:experiments}

This section discusses hyperparameters and additional details of the conducted experiments that could not be provided in the main text due to space limitations. For all experiments except the \textit{teach my agent} benchmark, we used RL algorithms from the \texttt{Stable Baselines 3} library \citep{stable-baselines}. For \textit{teach my agent}, we use the \sac implementation provided with the benchmark.

\subsection{Algorithm Hyperparameters}

\begin{table*}[t]
	\begin{center}
		\begin{small}
			\textsc{
				\begin{tabular}{lccccccccccr}
					\toprule
					& \multicolumn{4}{c}{\sprl} & & \multicolumn{2}{c}{\currot} & & \multicolumn{2}{c}{\gradient} \\
					\cmidrule{2-5} \cmidrule{7-8} \cmidrule{10-11}
					Env. & $\delta$ & $\epsilon$ & $\cvec{\sigma}_{\text{lb}}$ & $D_{\text{KL}_{\text{lb}}}$ & & $\delta$ & $\epsilon$ & & $\delta$ & $\epsilon$ \\
					\midrule
					Sparse Goal-Reaching & $0.6$ & $.25$ & - & - & & $0.8$ & $1.2$ & & $0.6$ & $0.05$ \\
					Point Mass & $4$ & $.25$ & $[.2\ .1875]$ & $8000$ & & $4$ & $0.7$ & & $3.0$ & $0.2$ \\
                    Unlock-Pickup & - & - & - & - & & $0.6$ & $3$ & & $0.6$ & $0.05$ \\
					Teach My Agent & - & - & - & - & & $180$ & $0.5 \vert 0.4$ & & $180$ & $0.05$ \\
					\bottomrule
				\end{tabular}
			}
		\end{small}
	\end{center}
	\caption{Hyperparameters of \sprl, \currot, and \gradient in the different learning environments. The $\epsilon$ parameter of \currot is computed according to the procedure described in appendix \ref{app:currot:experiments}. We do not provide \textit{teach my agent} parameters for \sprl as we rely on the results reported by \citep{romac2021teachmyagent}. We also do not evaluate \sprl in the unlock-pickup environment since \sprl is designed for continuous context spaces.}
	\label{tab:currot:interp_parameters}
	\vskip -0.1in
\end{table*}
\begin{table*}[b!]
	\begin{center}
		\begin{small}
			\textsc{
				\begin{tabular}{lcccccccccccr}
					\toprule
					& \multicolumn{3}{c}{\alpgmm} & & \multicolumn{3}{c}{\goalgan} & & \multicolumn{2}{c}{\acl} \\
					\cmidrule{2-4} \cmidrule{6-8} \cmidrule{10-11}
					Env. & $p_{\text{rand}}$ & $n_{\text{rollout}}$ & $s_{\text{buffer}}$ & & $\delta_{\text{noise}}$ & $n_{\text{rollout}}$ & $p_{\text{success}}$ & & $\eta$ & $\epsilon$ \\
					\midrule
					Sparse Goal-Reaching & $.2$ & $200$ & $500$ & & $.1$ & $200$ & $.2$ & & $0.05$ & $0.2$ \\
					Point Mass & $.1$ & $100$ & $500$ & & $.1$ & $200$ & $.2$ & & $0.025$ & $0.2$ \\
                    Unlock-Pickup & - & - & - & & - & - & - & & $0.025$ & $0.1$ \\
					\bottomrule
				\end{tabular}
			}
   \vspace{10pt}

			\textsc{
				\begin{tabular}{lccccccr}
					\toprule
					& \multicolumn{3}{c}{\plr} & & \multicolumn{3}{c}{\vds} \\
					\cmidrule{2-4} \cmidrule{6-8}
					Env. & $\rho$ & $\beta$ & $p$ & & lr & $n_{\text{ep}}$ & $n_{\text{batch}}$ \\
					\midrule
					Sparse Goal-Reaching & $.45$ & $.15$ & $.55$ & & $5 {\times} 10^{-4}$ & $10$ & $80$ \\
					Point Mass & $.15$ & $.45$ & $.85$ & &  $10^{-3}$ & $3$ & $20$ \\
                    Unlock-Pickup & $.45$ & $.45$ & $.55$ & & $10^{-3}$ & $5$ & $20$ \\ 
					\bottomrule
				\end{tabular}
			}
		\end{small}
	\end{center}
	\caption{Hyperparameters of the investigated baseline algorithms in the different learning environments, as described in Appendix \ref{app:currot:experiments}.}
	\label{table:currot:baseline_parameters}
\end{table*}

The main parameters of \sprl, \currot, and \gradient all factor into one parameter $\delta$ corresponding to the performance constraint and one parameter $\epsilon$ controlling the interpolation speed. We did not perform an extensive hyperparameter search for these parameters but used their interpretability to select appropriate parameter regions to search in. The performance parameter $\delta$ was chosen by evaluating values around $50\%$ of the maximum reward. This approach resulted in a search over $\delta \in \{3, 4, 5\}$ for the point-mass environment and  $\delta \in \{0.4, 0.6, 0.8\}$ for the sparse goal-reaching and unlock-pickup environment. For the \textit{teach my agent} experiments, we evaluated $\delta \in \{140, 160, 180\}$ for \currot and \gradient. We did not evaluate \sprl in the \textit{teach my agent} experiment since we took the results from \citet{romac2021teachmyagent}. We evaluated \gradient for $\epsilon \in [0.05, 0.1, 0.2]$. For \sprl, we initialized $\epsilon$ with a value of $0.05$ used in the initial experiments by Klink et al. However, we realized that larger values slightly improved performance. For \currot, the value of $\epsilon$ depends on the magnitude of the distances $d$ and hence changes per experiment. \rebuttal{In the conducted experiments, we set the parameter $\epsilon$ to around $5\%$ of the maximum distance between any two points in the context space, also evaluating a slightly larger and smaller value. However, we refer to Appendix \ref{app:currot:high-dim} for a detailed discussion of how to chose $\epsilon$ particularly when dealing with higher dimensional context spaces.} When targeting narrow target distributions, Klink et al. introduce a lower bound on the standard deviation $\cvec{\sigma}_{\text{lb}}$ of the context distribution of \sprl. This lower bound needs to be respected until the KL divergence w.r.t. $\mu(\svec{c})$ falls below a threshold $D_{\text{KL}}$, as otherwise, the variance of the context distribution may collapse too early, causing the KL divergence constraint on subsequent distributions to only allow for minimal changes to the context distribution. This detail again highlights the benefit of Wasserstein distances, as they are not subject to such subtleties due to their reliance on a chosen metric. Table \ref{tab:currot:interp_parameters} shows the parameters of \currot, \gradient, and \sprl for the different environments. \\
For \alpgmm, the relevant hyperparameters are the percentage of random samples drawn from the context space $p_{\text{rand}}$,
the number of completed learning episodes between the update of the context distribution $n_{\text{rollout}}$, and the
maximum buffer size of past trajectories to keep $s_{\text{buffer}}$. Similar to \citet{klink2021probabilistic}, we chose them by a grid-search over $(p_{\text{rand}}, n_{\text{rollout}}, s_{\text{buffer}}) \in \{0.1, 0.2, 0.3\} \times \{50, 100, 200\} \times \{500, 1000, 2000\}$. \\
For \goalgan, we tuned the amount of random noise that is added on top of each sample $\delta_{\text{noise}}$, the
number of policy rollouts between the update of the context distribution $n_{\text{rollout}}$ as well as the
percentage of samples drawn from the success buffer $p_{\text{success}}$ via a grid search over $(\delta_{\text{noise}}, n_{\text{rollout}}, p_{\text{success}}) \in \{0.025, 0.05, 0.1\} \times \{50, 100, 200\} \times \{0.1, 0.2, 0.3\}$.\\
For \acl, the continuous context spaces of the environments need to be discretized, as the algorithm is formulated as a bandit problem. The Exp3.S bandit algorithm that ultimately realizes the curriculum requires two hyperparameters to be chosen: the scale factor for updating the arm probabilities $\eta$ and the $\epsilon$ parameter of the $\epsilon$-greedy exploration strategy. We combine \acl with the absolute learning progress (ALP) metric also used in \alpgmm and conducted a hyperparameter search over $(\eta, \epsilon) \in \{0.05, 0.1, 0.2\} \times \{0.01, 0.025, 0.05\}$. Hence, contrasting \acl and \alpgmm sheds light on the importance of exploiting the continuity of the context space. For \acl, the absolute learning progress in a context $\svec{c}$ can be estimated by keeping track of the last reward obtained in the bin of $\svec{c}$ (note that we discretize the context space) and then computing the absolute difference between the return obtained from the current policy execution and the stored last reward. We had numerical issues when implementing the \acl algorithm by \citet{graves2017automated} due to the normalization of the ALPs via quantiles. Consequently, we normalized via the maximum and minimum ALP seen over the entire history of tasks. \\
For \plr, the staleness coefficient $\rho$, the score temperature $\beta$, and the replay probability $p$ need to be chosen. We did a grid search over $(\rho, \beta, p) \in \{0.15, 0.3, 0.45\} \times \{0.15, 0.3, 0.45\} \times \{0.55, 0.7, 0.85\}$ and chose the best configuration for each environment. \\
For \vds, the parameters for the training of the $Q$-function ensemble, i.e., the learning rate $\text{lr}$, the number of epochs $n_{\text{ep}}$ and the number of mini-batches $n_{\text{batch}}$, need to be chosen. Just as for \plr, we conducted a grid search over $(\text{lr}, n_{\text{ep}}, n_{\text{batch}}) \in \{10^{-4}, 5 {\times} 10^{-4}, 10^{-3}\} \times \{3, 5, 10\} \times \{20, 40, 80\}$.
The parameters of all employed baselines are given in Table \ref{table:currot:baseline_parameters}. We now continue with the description of experimental details for each environment.

\subsection{E-Maze Environment}
\label{app:currot:experiments-emaze}

\begin{figure}[t]
    \centering
    \includegraphics{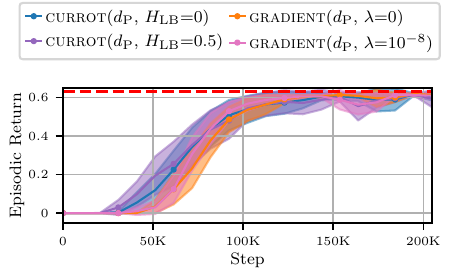}
    \caption{Expected return on the target task distribution $\mu(c)$ in the E-Maze environment achieved by \currot and \gradient under varying entropy regularizations for the current performance-based distance $d_{\text{P}}$. The shaded area corresponds to two times the standard error (computed from $20$ seeds). The red dotted line represents the maximum possible reward achievable on $\mu(c)$.}
    \label{fig:currot:emaze-cur-perf}
\end{figure}

\begin{figure}[b!]
    \centering
    \includegraphics{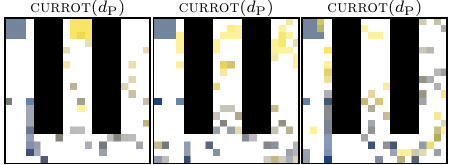}
    \includegraphics{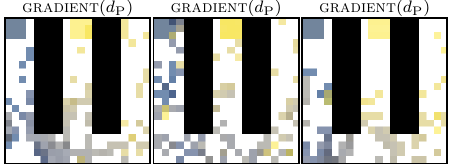}
    \caption{Interpolations generated by \currot and \gradient in different runs for the current performance-based distance $d_{\text{P}}(c_1, c_2)$. Brighter colors indicate later iterations.}
    \label{fig:currot:emaze-cur-perf-interpolations}
\end{figure}

The $xy$-coordinates of the representatives 
$$\svec{r}(c) {=} [x,\ y,\ z]^T \in \mathbb{R}^3$$ 
of a context $c$ form a grid on ${[-1,1] \times [-1, 1]}$ and, as mentioned in the main article, $z{=}200$ for walls and $z{=}0$ for all other cells. The four actions $\{\text{up}, \text{down}, \text{left}, \text{right}\}$ lead to a transition to the corresponding neighboring cell with a probability of $0.9$, if the neighboring cell has the same height, and $0$ if not. Upon reaching the desired state (controlled by the context $c$), the agent observes a reward of value one, and the episode terminates. In this environment, we use \ppo with $\lambda=0.99$ and all other parameters left to the implementation defaults of the \texttt{Stable Baselines 3} library. \\
For solving objectives (\ref{eq:currot:currot}) and (\ref{eq:currot:gradient}), we make use of the \texttt{linprog} function from the \texttt{SciPy} library \cite{2020SciPy-NMeth}. \\
\textbf{Current Agent Performance as a Distance:} In the main text, we have investigated the pseudo-distance 
\begin{align}
    d_{\text{P}^*}(c_1, c_2) = | J(\pi^*, c_1) - J(\pi^*, c_2) | 
\end{align}
that defines the similarity of contexts based on the absolute performance difference of the optimal policy in the contexts $c_1$ and $c_2$. While $d_{\text{P}^*}$ only performed slightly worse than the more informed distance $d_{\text{S}}$ for \gradient, it could only provide meaningful performance for \currot if combined with entropy regularization. However, \citet{huang2022curriculum} also investigated a pseudo-distance function that computes the similarity of two contexts based on the \emph{current} policy $\pi$ 
\begin{align}
    d_{\text{P}}(c_1, c_2) = | J(\pi, c_1) - J(\pi, c_2) |,
\end{align}
leading to a distance function that changes in each iteration. As we show in Figure \ref{fig:currot:emaze-cur-perf}, this distance, while still leading to slower learning for \currot compared to $d_{\text{S}}$, leads to stable learning across different levels of entropy regularization without any prior environment knowledge. Figure \ref{fig:currot:emaze-cur-perf-interpolations} shows multiple curricula that have been generated by \currot and \gradient. Particularly for \currot, we can see fairly diverse curricula, which sometimes target all three corridors at once (top middle) and sometimes even back track out of the right-most corridor into the remaining two (top right). We see the good performance and the diverse behavior as indicators for the potential of general purpose distance metrics that encode some form of implicit exploration, calling for future investigations to better understand their mechanics. Furthermore, computational aspects arise with the use of such metrics, since for the case of $d_{\text{P}}$, robust and efficient versions for estimating $J(\pi, c)$ need to be devised.
\\
\textbf{Entropy-Regularized \currot and \gradient:} As discussed in Section \ref{sec:currot:e-maze}, we benchmark versions of \gradient and \currot in which we introduce different forms of entropy regularization. For \gradient, we recreate the implementation by \citet{huang2022curriculum} by using optimal transport formulations that regularize the entropy of the transport plan $\phi$ \citep{benamou2015iterative,cuturi2013sinkhorn}
\begin{align}
    \mathcal{W}_{p,\lambda}(p_1, p_2) &{=} \left( \inf_{\phi \in \Phi(p_1, p_2)}  \mathbb{E}_{\phi} \left[d(\svec{c}_1, \svec{c}_2)^p \right] - \lambda H(\phi) \right)^{1 / p},
\end{align} \\
with the constraint set $\Phi(p_1,p_2)$ defined as in Section \ref{sec:currot:ot} and the entropy $H(p)$ of a distribution $p$ over a sample space $\mathcal{X}$ defined as $H(p) = -\int_{\svec{x} \in \mathcal{X}} p(\svec{x}) \log(p(\svec{x}))$. Note that \citet{huang2022curriculum} chose these formulations for computational speed rather than curriculum performance. This formulation allows for a straightforward adaptation of the \gradient objective to incorporate entropy-regularization 
\begin{align}
    \max_{\alpha \in [0,1]}&\ \alpha \quad \text{s.t.}\ J(\pi, p_{\alpha, \lambda}) \geq \delta \\
    p_{\alpha, \lambda}(c) =& \argmin_p\  \alpha \mathcal{W}_{2,\lambda}(p, \mu) + (1 - \alpha) \mathcal{W}_{2,\lambda}(p, p_0).
\end{align}
For the \currot algorithm, we choose a more direct form of regularization and directly constrain the entropy of the interpolating distribution $p$
\begin{align}
    \min_{p}&\ \mathcal{W}_2(p, \mu) \\
    \text{s.t.}&\ p(\mathcal{V}(\pi,\delta)) = 1 \quad H(p) \geq H_{\text{LB}} \nonumber.
\end{align}
The above entropy regularized objectives are not linear programs anymore, and we hence solve the (convex) objectives with the \texttt{CVXPY} library \cite{diamond2016cvxpy}.

\subsection{Unlock-Pickup Environment}
\label{app:currot:experiments-unlockpickup}

We use the Unlock-Pickup environment from the Minigrid library \citep{minigrid}. We do not change the behavior of the environment and only remove the additional discounting that occurs within the environment, as the environment does not reveal the current timestep to the agent, which, combined with an internally discounted reward, leads to non-Markovian behavior. As stated in the main article, the context $c$ controls the initial state of the environment by specifying the position of the agent, key, and box as well as the position and state of the door (i.e., open or closed). We use the \dqn algorithm since the extremely sparse nature of the environment favors RL algorithms with a replay buffer. Compared to the default parameters of the \dqn algorithm, we only increase the exploration rate from $0.05$ to $0.1$ and also increase the batch size to $256$. We train the $Q$-network every fourth step, updating the target network with a Polyak update with $\tau=0.005$ in each step. \\
The $Q$-network is realized by encoding the image observation with a convolutional neural network with three convolutions of kernel size $(2,2)$, ReLU activations after each convolution, and a max-pool operation with kernel size $(2,2)$ after the first convolution and ReLU operation. We do not use information about the agent orientation or the textual task description, as both are not strictly necessary for our environment. The convolutional network has $32$-dimensional hidden layers. The output of the convolutional encoder is $64$-dimensional, which is then further processed by two fully connected layers with $64$ dimensions and ReLU activations before being reduced to the $Q$-values for the seven actions available in the environment. \\
As briefly mentioned in the main article, the target distribution $\mu(c)$ is a uniform distribution over all those contexts in which the agent is in the left room with a closed door and does not hold the key. The initial state distribution contains one context for each box position in the right room in which the agent is positioned directly next to the box. \\
\textbf{Distance Function} As discussed in Section \ref{sec:currot:unlockpickup}, a context $c$ controls the starting state of the environment, which is defined by
\begin{itemize}
    \item the agent position $\text{ap}: \mathcal{C} \mapsto [1,9] \times [1,4]$
    \item the key position $\text{kp}: \mathcal{C} \mapsto [1,9] \times [1,4]$
    \item the box position $\text{bp}: \mathcal{C} \mapsto [6,9] \times [1,4]$
    \item the position of the door in the wall $\text{dp}: \mathcal{C} \mapsto [1,4]$
    \item the state of the door $\text{ds}: \mathcal{C} \mapsto \{\text{open},\text{closed}\}$.
\end{itemize}
The images of the individual functions that access the state information of a context are motivated by the two rooms $R_1 = [1,4] \times [1,4]$ and $R_2 = [6,9] \times [1,4]$ that make up the environment. Consequently, the agent and the key can be placed in both rooms, whereas the box can only be placed in $R_2$. The wall that separates the rooms occupies tiles in $W(c) = \{(5,y)\ | \ y {\in} [1,4], y {\neq} \text{dp}(c) \} $. Due to this wall, we restrict the context space $\mathcal{C}$ such that it does not contain contexts in which the agent or key is located in the wall, i.e., $\text{ap}(c) \notin W(c)$ and $\text{kp}(c) \notin W(c)$. Additionally, we only allow placing the agent and key in $R_2$ if the door is open. Formally, this requires $\text{ap}(c) {\geq} 4 \Rightarrow \text{ds}(c) {=} \text{open}$ and $\text{kp}(c) {\geq} 4 \Rightarrow \text{ds}(c) {=} \text{open}$. Finally, neither key nor agent can be at the same position as the box, i.e., $\text{ap}(c) \neq \text{bp}(c)$ and $\text{kp}(c) \neq \text{bp}(c)$. With these restrictions, we arrive at the $81.920$ individual contexts mentioned in Section \ref{sec:currot:unlockpickup}.\\
Note that the distance function between contexts reasons both about state changes that can be achieved in an episode, such as moving between agent positions, and ones that can't, such as moving the box. Moving boxes is impossible since the episode terminates successfully when the agent picks up the box. Hence, a distance function that is purely based on state transitions would neglect certain similarities between contexts in this environment.\\
We define the distance function $d_{\text{base}}(c_1, c_2)$ function via representatives $r(c)$, i.e.
\begin{align}
    d(c_1, c_2) {=} \begin{cases}
        d_{\text{base}}(c_1, r(c_1)) {+} d_{\text{base}}(r(c_1), r(c_2)) \\
        \quad {+} d_{\text{base}}(r(c_2), c_2),\ \text{if } \text{ds}(c_1) {\neq} \text{ds}(c_2) \\
        d_{\text{base}}(c_1, c_2),\ \text{else.}
    \end{cases}
\end{align}
Such distances are also known as highway distances \cite{baikousi2011similarity}. The mapping $r: \mathcal{C} \mapsto \mathcal{C}$ from a context $c$ to its representative $r(c)$ ensures that the agent is standing right in front of the open door with the key in its hand, i.e., $\text{ds}(r(c)){=}\text{open}$, and $\text{ap}(r(c)){=}\text{kp}(r(c)){=}[4,\text{dp}(c)]$, while ensuring that $\text{dp}(r(c)){=}\text{dp}(c)$ and $\text{bp}(r(c)){=}\text{bp}(c)$. \\
The base distance $d_{\text{base}}(c_1, c_2)$ encodes the cost of moving both key and agent from their positions in $c_1$ to those in $c_2$ (via $d_{\text{ka}}$) as well as the cost of equalizing the box positions between the contexts (via the L1 distance)
\begin{align}
    d_{\text{base}}(c_1,c_2) = \begin{cases}
    d_{\text{ka}}(c_1, c_2) + \| \text{bp}(c_1) - \text{bp}(c_2) \|_1,\ \\
    \quad \text{if }  \text{dp}(c_1) {=} \text{dp}(c_2) \\ 
    \infty,\ \text{else.}
    \end{cases}
\end{align}
We see that we render contexts with different door positions incomparable to ease the definition of the distance function. The key-agent distance is defined on top of an object distance $d_{\text{obj,dp}}$ that is conditioned on a door position $\text{dp}$
\begin{align}
    d_{\text{ka}}(c_1,c_2) = \begin{cases}
    d_{\text{obj,dp}(c1)}(\text{ap}(c_1), \text{ap}(c_2)),\ \text{if }  \text{kp}(c_1) {=} \text{kp}(c_2) \\ 
    d_{\text{obj,dp}(c1)}(\text{ap}(c_1), \text{kp}(c_1)) \\
    \quad + d_{\text{obj,dp}(c1)}(\text{kp}(c_1), \text{kp}(c_2)) \\
    \quad + d_{\text{obj,dp}(c1)}(\text{ap}(c_2), \text{kp}(c_2)),\ \text{else.}
    \end{cases}
\end{align}
Note that we can simply take $\text{dp}(c_1)$ since we know that $\text{dp}(c_1) {=} \text{dp}(c_2)$. The object distance is defined as the L1 distance between the two objects if they are in the same room and incorporates the detour caused by passing through the door in the wall if not
\begin{align}
    d_{\text{obj,dp}}(\svec{p}_1, \svec{p}_2) = \begin{cases}
    \|\svec{p}_1 - \svec{p}_2\|_1,\ \text{if }  p_{1,0} {\leq} 4 \Leftrightarrow p_{2,0} {\leq} 4 \\ 
     \|\svec{p}_1 - [5,\text{dp}]\|_1 + \|[5,\text{dp}] - \svec{p}_2\|_1,\ \text{else.}
    \end{cases}
\end{align}
We ensured that the resulting distance $d(c_1,c_2)$ fulfills all axioms of a valid distance function, i.e. $d(c_1,c_2) \geq 0$, $d(c_1,c_2) {=} 0 \Leftrightarrow c_1 {=} c_2$, $d(c_1,c_2) = d(c_2,c_1)$, and $d(c_1,c_3) \leq d(c_1,c_2) + d(c_2,c_3)$ via brute-force computations. Note that the in-comparability of contexts with different door positions effectively splits the context space into four disjoint sets (for the four different door positions) that cannot be compared. Hence, we must only ensure these axioms within the four disjoint sets separately. %
\begin{figure}
    \centering
    \includegraphics{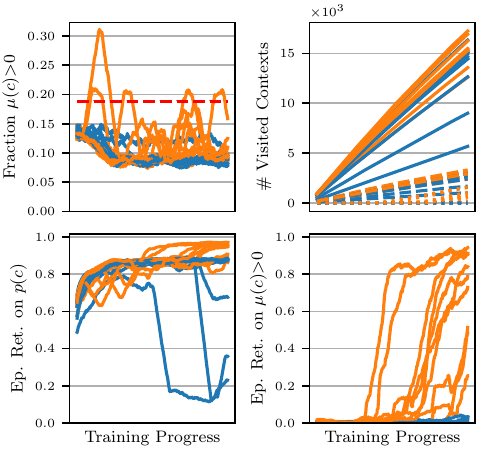}
    \caption{Statistics of the \plr curricula in the unlock-pickup environment over training progress. The top left plot shows the fraction of contexts sampled by \plr that are also sampled by the target context distribution $\mu(c)$. The red dashed line indicates the fraction of target samples generated by a random curriculum. The top right plot shows the number of unique contexts (solid lines), unique target contexts (dashed lines), and unique solved target contexts (dotted lines) sampled by \plr at least once. The bottom left plot indicates the performance on the \plr curriculum. The performance in those contexts of the curriculum, which are also sampled by the target context distribution $\mu(c)$ (i.e., on the fraction indicated in the top left), are shown in the bottom right.}
    \label{fig:currot:plr_unlock_pickup}
\end{figure} \\
\noindent \textbf{PLR Performance:} As mentioned in Section \ref{sec:currot:unlockpickup}, Figure \ref{fig:currot:plr_unlock_pickup} shows statistics of the \plr curricula. We can see that throughout most \plr curricula, the chance of sampling a target context stays relatively constant, even though the number of distinct sampled contexts and the number of distinct sampled target contexts continuously grows.  We also see that the agent receives a positive learning signal on $p(c)$ in all runs of \plr. Additionally, we see that the prioritization by \plr suppresses contexts from $\mu(c)$ since a purely random curriculum would sample a target context $18.75\%$ of the time. In about half of the runs (orange lines), the agent learned to solve some of the target tasks, although this fraction is rather low (there are $15.360$ target tasks). Interestingly, this increase in proficiency on tasks from $\mu(c)$ does not go hand-in-hand with a consistently increased sampling rate of target tasks. However, as we see in Figure \ref{fig:currot:plr_contexts} there seems to be a tendency of \plr runs that are more successful on $\mu(c)$ to sample more contexts in which the agent is located in the left room at the beginning of the episode. Generally speaking, Figures \ref{fig:currot:plr_unlock_pickup} and \ref{fig:currot:plr_contexts} show that \plr prioritized specific contexts over others. However, either due to the missing notion of a target distribution or the dependence of \plr on the agent's internal value function (which may be biased and incorrect), the generated curricula did not consistently progress to the most challenging, long-sequence tasks encoded by $\mu(c)$.
\begin{figure}
    \centering
    \includegraphics{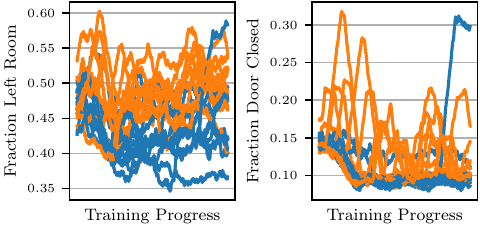}
    \caption{Fraction of contexts in the \plr curricula in which the agent is placed in the left room (left) and in which the door is closed (right) at the start of the episode. A closed door implies that the agent is located in the left room, hence a more strict condition. Note that the color coding corresponds to the one in Figure \ref{fig:currot:plr_unlock_pickup}, indicating runs with high- (orange) and low performance (blue) on $p(c)$.}
    \label{fig:currot:plr_contexts}
\end{figure}

\subsection{Point-Mass Environment}
\label{app:currot:experiments-point-mass}

\begin{figure*}[t]
    \centering
    \begin{subfigure}[b]{0.19\textwidth}
		\centering
		\includegraphics{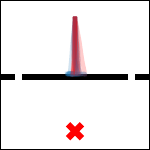}
		\caption{Default}
    \end{subfigure}
    \begin{subfigure}[b]{0.19\textwidth}
		\centering
		\includegraphics{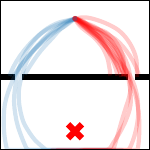}
		\caption{Random}
    \end{subfigure}
    \begin{subfigure}[b]{0.19\textwidth}
		\centering
		\includegraphics{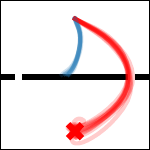}
		\caption{\sprl}
    \end{subfigure}
    \begin{subfigure}[b]{0.19\textwidth}
		\centering
		\includegraphics{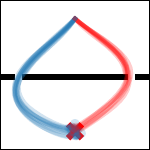}
		\caption{\currot}
    \end{subfigure}
    \begin{subfigure}[b]{0.19\textwidth}
		\centering
		\includegraphics{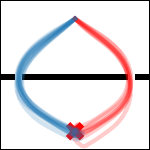}
		\caption{\gradient}
    \end{subfigure}
        \begin{subfigure}[b]{0.19\textwidth}
		\centering
		\includegraphics{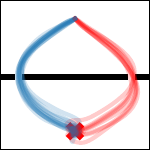}
		\caption{\alpgmm}
    \end{subfigure}
        \begin{subfigure}[b]{0.19\textwidth}
		\centering
		\includegraphics{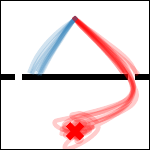}
		\caption{\goalgan}
    \end{subfigure}
    \begin{subfigure}[b]{0.19\textwidth}
		\centering
		\includegraphics{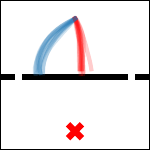}
		\caption{\acl}
    \end{subfigure}
    \begin{subfigure}[b]{0.19\textwidth}
		\centering
		\includegraphics{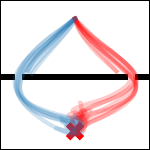}
		\caption{\plr}
    \end{subfigure}
    \begin{subfigure}[b]{0.19\textwidth}
		\centering
		\includegraphics{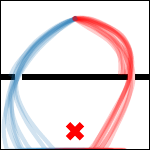}
		\caption{\vds}
    \end{subfigure}
    \caption{Final trajectories generated by the different investigated curricula in the point mass environment. The color encodes the context: Blue represents gates positioned at the left and red at the right.}
    \label{fig:currot:point_mass_trajectories}
\end{figure*}

\begin{figure*}[b!]
\centering
\begin{subfigure}[c]{0.3\textwidth}
\centering
\includegraphics{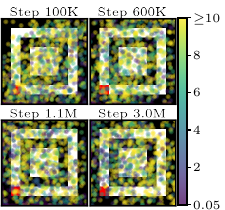}
\subcaption{\sprl Curriculum (SGR)}
\label{fig:currot:sprl_maze_curriculum}
\end{subfigure}
\begin{subfigure}[c]{0.39\textwidth}
\centering
\includegraphics{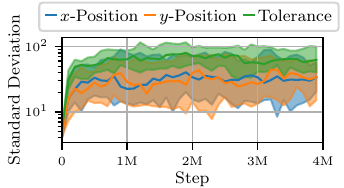}
\subcaption{\sprl Sampling Distribution Stds. (SGR)}
\label{fig:currot:sprl_maze_stds}
\end{subfigure}
\begin{subfigure}[c]{0.3\textwidth}
\centering
\includegraphics{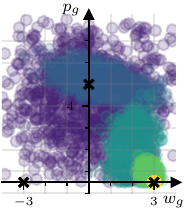}
\subcaption{\sprl Curriculum (Point Mass)}
\label{fig:currot:sprl_point_mass_curriculum}
\end{subfigure}
\caption{a) Visualization of the sampling distribution of \sprl in the sparse goal-reaching (SGR) task. The color of the dots encodes the tolerance of the corresponding contexts, and the position represents the goal to be reached under that tolerance. The walls are shown in black, and the red area visualizes the starting area of the agent. b) $10$-, $50$- and $90$-percentile of the standard deviation of \sprl's sampling distribution on the sparse goal-reaching task. The statistics have been computed from $20$ seeds. c) Sampling distribution of \sprl in the point mass environment for a given seed. The color indicates the iteration, where brighter colors correspond to later iterations.}
\label{fig:currot:sprl_qualitative_results}
\end{figure*}

The environment setup is the same as the one investigated by \citet{klink2020self,klink2021probabilistic} with the only difference in the target context distributions, which is now defined as a Gaussian mixture
\begin{align*}
    \mu(\svec{c}) &= \frac{1}{2} \mathcal{N}\left(\svec{c}_1, 10^{-4} \svec{I} \right) + \frac{1}{2} \mathcal{N} \left(\svec{c}_2, 10^{-4} \svec{I} \right)\\
    \svec{c}_1 &= [-3\ 0.5]^T, \svec{c}_2 = [3\ 0.5]^T.
\end{align*}
In this environment, we use \ppo with $4.096$ steps per policy update, a batch size of $128$, and $\lambda{=}0.99$. All other parameters are left to the implementation defaults of the \texttt{Stable Baselines 3} implementation.\\
Figure \ref{fig:currot:point_mass_trajectories} shows trajectories generated by agents trained with different curricula in the point-mass environment. We see that directly learning on the two target tasks (Default) prevents the agent from finding the gates in the wall to pass through. Consequently, the agent minimizes the distance to the goal by moving right in front of the wall (but not crashing into it) to accumulate reward over time. We see that random learning indeed generates meaningful behavior. This behavior is, however, not precise enough to pass reliably through the wall. As mentioned in the main article, \sprl only learns to pass through one of the gates, as its uni-modal Gaussian distribution can only encode one of the modes of $\mu(\svec{c})$ (see Figure \ref{fig:currot:sprl_qualitative_results} for a visualization). \currot and \gradient learn policies that can pass through both gates reliably, showing that the gradual interpolation towards both target tasks allowed the agent to learn both. \alpgmm and \plr also learn good policies. The generated trajectories are, however, not as precise as the ones learned with \currot and \gradient and sometimes only solve one of the two tasks reliably. \acl, \goalgan, and \vds partly create meaningful behavior. However, this behavior is unreliable, leading to low returns due to the agent frequently crashing into the wall.

\subsection{Sparse Goal-Reaching Environment}
\label{app:currot:experiments-sgr}

For the sparse goal-reaching task, the goal can be chosen within $[-9, 9] \times [-9, 9]$, and the allowed tolerance can be chosen from $[0.05, 18]$. Hence, the context space is a three-dimensional cube $\mathcal{C} = [-9, 9] \times [-9, 9] \times [0.05, 18]$. The actually reachable space of positions (and with that goals) is a subset of $[-7, 7] \times [-7, 7]$ due to the ``hole'' caused by the inner walls of the environment. The target context distribution is a uniform distribution over tasks with a tolerance of $0.05$
\begin{align*}
    \mu(\svec{c}) \propto \begin{cases}
    1,\ \text{if } c_3 = 0.05, \\
    0,\ \text{else.}
    \end{cases}
\end{align*}
The state $\svec{s}$ of the environment is given by the agent's $x$- and $y$-position. The reward is sparse, only rewarding the agent if the goal is reached. A goal is considered reached if the Euclidean distance between the goal and position of the point mass falls below the tolerance
\begin{align*}
    \| \svec{s} - [\svec{c}_1\ \svec{c}_2]^T \|_2 \leq c_3.
\end{align*}
The two-dimensional action of the agent corresponds to its displacement in the $x-$ and $y-$ direction. The action is clipped such that the Euclidean displacement per step is no larger than $0.3$. \\
Given the sparse reward of the task, we again use an RL algorithm that utilizes a replay buffer. Since the actions are continuous in this environment, we use \sac instead of \dqn. Compared to the default algorithm parameters of \texttt{Stable Baselines 3}, we only changed the policy update frequency to $5$ environment steps, increased the batch size to $512$, and reduced the buffer size to $200.000$ steps. \\
Figure \ref{fig:currot:sprl_qualitative_results} visualizes the behavior of \sprl in the sparse goal-reaching (SGR). We see that for the SGR environment, \sprl increases the variance of the Gaussian context distribution to assign probability density to the target contexts while fulfilling the expected performance constraint by encoding trivial tasks with high tolerance (Figures \ref{fig:currot:sprl_maze_curriculum} and \ref{fig:currot:sprl_maze_stds}). The inferior performance of an agent trained with \sprl compared to one trained with a random curriculum shows that the Gaussian approximation to a uniform distribution is a poor choice for this environment. While it may be possible to find other parametric distributions that are better suited to the particular problem, \currot flexibly adapts the shape of the distribution without requiring any prior choices. \\
\begin{figure}[t]
    \centering
    \includegraphics{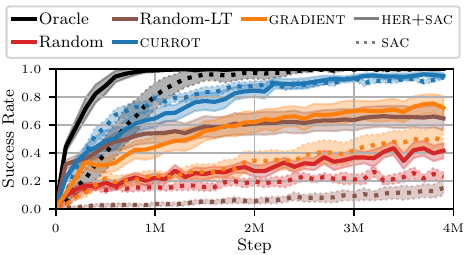}
    \caption{\revision{Comparison of Hindsight Experience Replay (\her, solid lines) and \sac (dotted lines). Across all curricula, pairing \her and \sac achieves similar or better final success rate compared to \sac alone. The final success rate improves the most when training on random tasks with the target tolerance of $0.05$ (\textbf{Random-LT}). When training on random tasks with randomized tolerance (\textbf{Random}), performance improvements are less pronounced. Mean and two-times standard error intervals are computed from $20$ seeds.}{}}
    \label{fig:currot:her_sgr}
    \vspace{-5pt}
\end{figure} \\ 
\revision{\textbf{Hindsight Experience Replay (\her)}: Given the success of \her for sparse-reward goal-reaching tasks, we evaluated its performance in our sparse goal-reaching environment. A difference to the environments evaluated by \citet{andrychowicz2017hindsight} is the varying tolerance encoded by the contexts $\mathbf{c} {\in} \mathbb{C} \subseteq \mathbb{R}^3$. \citet{andrychowicz2017hindsight} assumed a fixed tolerance for their investigations of \her. We consequently train \her by uniformly sampling $\mathcal{C}$, corresponding to the Random strategy in Figure \ref{fig:currot:o-maze-performance}, and sampling from $\mu(\mathbf{c})$, i.e., only sampling high-precision tasks. We refer to the latter sampling strategy as \textbf{Random-LT}, where LT is short for low tolerance. \her only influences experience replay and can be easily combined with arbitrary task sampling strategies. Figure \ref{fig:currot:her_sgr} shows the results of training \her with the aforementioned task-sampling strategies and in combination with \gradient and \currot. We used the \her implementation in the \texttt{Stable Baselines 3} library \citep{stable-baselines} with the \textit{future} strategy. We tuned the number of additional goals to maximize \her's performance, finding that $k{=}2$ additional goals for each real goal delivered the best results. Looking at Figure \ref{fig:currot:her_sgr}, we see that \her is well-compatible with all curricula, either matching or improving upon the success rate of \sac alone. \her drastically improves the performance when directly sampling high-precision tasks of $\mu(\mathbf{c})$. Training on random tasks of $\mathcal{C}$ and with \gradient benefit from \her, whereas the performance of \currot does not improve with the replay of hindsight goals. Finally, when training only on the feasible tasks of $\mu(\svec{c})$ (Oracle), \her significantly improves learning speed. The results indicate that for this task, \her's implicit curriculum has a somewhat orthogonal effect than the explicit curricula realized by the different investigated sampling strategies.}{}

\subsection{Teach My Agent}

As mentioned in the main text, we used the environment and \sac learning agent implementation provided by \citet{romac2021teachmyagent}. We only interfaced \currot and \gradient to the setup they provided, allowing us to reuse the baseline evaluations provided by \citet{romac2021teachmyagent}. The two settings (\emph{mostly infeasible} and \emph{mostly trivial}) differ in the boundaries of their respective context spaces. The \emph{mostly infeasible} setting encodes tasks with a stump height in $[0,9]$ and -spacing in $[0,6]$. The \emph{mostly trivial} setting keeps the same boundaries for the stump spacing while encoding stumps with a height in $[-3,3]$. Since a stump with negative height is considered not present, half of the context space of the \emph{mostly trivial} setting does not encode any obstacles for the bipedal walker to master. The initial- and target context distribution $\mu(\svec{c})$ is uniform over the respective context space $\mathcal{C}$ for both settings.

\section{Higher Dimensional Context Experiments}
\label{app:currot:high-dim}

\begin{figure*}
    \centering
    \includegraphics{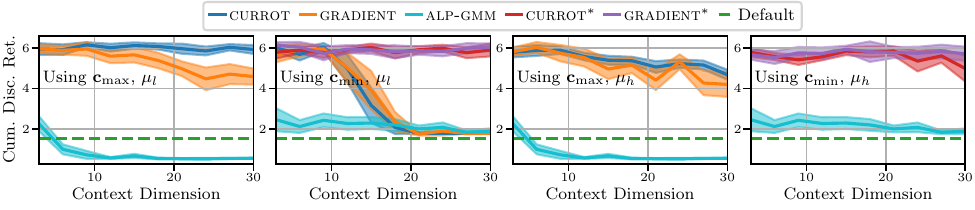}
    \caption{\rebuttal{Performance of \currot and \gradient in high-dimensional context space versions of the point mass environment. The two left plots show the final agent performance when training for the low-entropy target distribution $\mu_l(\svec{c})$ (Eq. \ref{eq:currot:point-mass-nd-l-target}) for different reductions $\svec{c}_{\text{max}}$ and $\svec{c}_{\text{min}}$. The two right plots shows the same results when training for the high-entropy target distribution $\mu_h(\svec{c})$. Note that the performance of \alpgmm is not affected by a change in target distribution since it generates the curriculum without this information. The green line indicates the average final performance of regular training on $\mu(\svec{c})$. Means (thick lines) and two-times standard errors (shaded areas) are computed from $20$ seeds. $\textsc{currot}^*$ and $\textsc{gradient}^*$ refer to versions of \currot and \gradient that use the less adversarial initial task distribution for the $\svec{c}_{\text{min}}$ reduction (please see Appendix \ref{app:currot:high-dim} for a description).}}
    \label{fig:currot:high_dim_point_mass}
\end{figure*}

\rebuttal{In addition to the low-dimensional context parametrizations of the tasks in the main article, we create a higher-dimensional version of the point-mass environment in which we essentially over-parameterize the environment. We do this by keeping the position of the gate $p_g \in [-4, 4]$ as a parameter but splitting the gate width into a left- and right width parameter $w_{g,l} \in [0.25, 4]$ and $w_{g,r} \in [0.25, 4]$. Note that we multiplied the range of two width parameters by a factor of $0.5$ compared to the regular point mass environment from the main article. The actual context for this environment consists of multiple instances of these three parameters, i.e.
\begin{align*}
    \svec{c} {=} \left[p_{g_1} \ldots p_{g_N}\, w_{g_1,l} \ldots \, w_{g_N,l}\, w_{g_1,r} \ldots \, w_{g_N,r} \right] \in \mathcal{C} {\subseteq} \mathbb{R}^{3N}.
\end{align*}
We instantiate the point-mass environment from this over-parameterized context using two different reductions
\begin{align}
    \svec{c}_{\text{min}} &= \left[ p_{g_{n^*}}\, \min_{n \in [1,N]} w_{g_n,l}\, \min_{n \in [1,N]} w_{g_n,r} \right] \label{eq:currot:context-reductions} \\
    \svec{c}_{\text{max}} &= \left[ p_{g_{n^*}}\, \max_{n \in [1,N]} w_{g_n,l}\, \max_{n \in [1,N]} w_{g_n,r} \right] \nonumber \\
    n^* &= \argmax_{n \in [1,N]} \left| p_{g_{n}} \right|. \nonumber
\end{align}
The only difference between the environment in Section \ref{sec:currot:point-mass} and the one investigated in this section is that we separately parameterize the width of the left- and right gate half. We chose these two reductions to highlight that not only the dimensionality of the context space $\mathcal{C}$ is important but also its underlying structure. When using $\svec{c}_{\text{min}}$, the chance of sampling tasks with a narrow gate far away from the center increases with $N$. For $\svec{c}_{\text{max}}$, the chance of sampling wide gates increases. Most importantly, the learning task does not get more complex with an increasing value of $N$ since the agent always faces the same learning task and observation space. We can hence be sure that observed performance drops are not due to an inherently more complex learning- or approximation task on the level of the RL agent but are due to the curriculum generation. We first investigate a narrow Gaussian mixture model as the target distribution
\begin{align}
    \mu_l(\svec{c}) &= \frac{1}{2} \mathcal{N}\left(\svec{c}_1, 10^{-4} \svec{I} \right) + \frac{1}{2} \mathcal{N} \left(\svec{c}_2, 10^{-4} \svec{I} \right) \label{eq:currot:point-mass-nd-l-target} \\
    \svec{c}_1 &= [\underbrace{-3 \ldots {-3}}_{N{-}\text{times}}\, \underbrace{0.25 \ldots 0.25}_{N{-}\text{times}}\, \underbrace{0.25 \ldots 0.25}_{N{-}\text{times}}] \\
    \svec{c}_2 &= [\underbrace{3 \ldots 3}_{N{-}\text{times}}\, \underbrace{0.25 \ldots 0.25}_{N{-}\text{times}}\, \underbrace{0.25 \ldots 0.25}_{N{-}\text{times}}]. \nonumber
\end{align} \\
We benchmarked \currot, \gradient, and \alpgmm in this task, keeping all algorithm parameters the same as in the point-mass environment from the main article and only adjusting the trust region parameter $\epsilon$ of \currot according to the rule described in Appendix \ref{app:currot:experiments} since the effective distances between points in $\mathcal{C}$ increase with $N$. Importantly, we always represent the curricula for \currot and \gradient using $100$ particles. Figure \ref{fig:currot:high_dim_point_mass} shows the obtained results.
As we see, \currot and \gradient generate good curricula even for high-dimensional context spaces when using the $\svec{c}_{\text{max}}$ reduction but fail for higher-dimensional scenarios when using $\svec{c}_{\text{min}}$. However, this failure does not arise from a failing interpolation but due to the increasing likeliness of sampling complex tasks under the initial uniform distribution over $\mathcal{C}$, leading to \currot and \gradient not reaching the performance threshold $\delta$ on the initial distribution $p_0(\svec{c})$. We first tested the feasible context search from the TeachMyAgent benchmark to remedy this issue. However, this search also failed since, just like for uniform noise, uninformed Gaussian noise increases the chance of sampling small gates for $\svec{c}_{\text{min}}$ in high dimensions. To benchmark the algorithms for the $\svec{c}_{\text{min}}$ reduction, we consequently generate an initial distribution with the same distribution of gate positions and -widths as for $N{=}1$, regardless of the choice of $N$. We do this by sampling contexts for $N{=}1$ and then projecting them to the required dimension by sampling appropriate random values for the remaining entries in $\svec{c}$. Starting from this initial distribution makes the agent proficient on $\mu(\svec{c})$ across all dimensions, as shown in Figure \ref{fig:currot:high_dim_point_mass} (we denote the resulting approaches as $\textsc{currot}^*$ and $\textsc{gradient}^*$). \\
We additionally investigate a setting where $\mu_h(\svec{c})$ encodes all high-dimensional contexts $\svec{c}$ that result in the same reduced target contexts $\svec{c}_1 {=} [-3\,0.25\,0.25]$ and $\svec{c}_2 {=} [3\,0.25\,0.25]$. When evaluating the \currot method in this scenario, we saw that our approximate optimization of Objective (\ref{eq:currot:currot-particle-opt}) via uniform samples in a half-sphere did not lead to good progression to the target samples. Adding samples along the direction $\svec{c}_{\mu,\phi(n)} {-} \svec{c}_{p,n}$ was enough to solve this issue in our approximate optimization and ensure good progression. While performing these experiments, we saw that our rule of choosing the $\epsilon$ parameter for \currot, i.e., setting it to $0.05$ of the maximum distance $d_{\text{max}}$ between any two contexts in $\mathcal{C}$, can prevent Objective (\ref{eq:currot:currot-particle-opt}) from sampling high-dimensional high-entropy distributions. This problem occurs if the Wasserstein distance between two particle-based representations $\hat{\mu}_1(\svec{c})$ and $\hat{\mu}_2(\svec{c})$ of the target distribution $\mu(\svec{c})$ is larger than $\epsilon {=} 0.05 d_{\text{max}}$. Consequently, we adapted our rule of choosing the trust region size for \currot to $\epsilon {=} \max(0.05 d_{\text{max}}, 1.2 \mathcal{W}_2(\hat{\mu}_1, \hat{\mu}_2))$. Figure \ref{fig:currot:high_dim_point_mass} shows the results for the high-entropy target distributions, where we again see that both \currot and \gradient can solve these tasks. \\
\noindent \textbf{Choosing Particles in Higher Dimensions:} The findings in this section provided a better understanding of the role of the number of particles to represent $\hat{p}(\svec{c})$ that we would like to summarize here. \\
For \currot and \gradient, the particles serve two objectives: Approximating the sampling- and target distribution and estimating the agent performance. By restricting the curriculum to the barycentric interpolation, \gradient can provide unbiased samples from the interpolation and the target distribution even when using a few particles in high dimensions. Consequently, the need for more particles in higher dimensions only arises from counteracting a potentially higher variance of the expected performance estimate. However, the effect of more noisy expected performance estimates can also be counteracted by smaller step sizes $\epsilon$ with which to advance the Barycentric interpolation. \\
For \currot, we saw that a small number of particles in combination with a too-small trust region $\epsilon$ can lead to biased sampling of the target distribution. However, we also saw that setting $\epsilon {=} \max(0.05 d_{\text{max}}, 1.2 \mathcal{W}_2(\hat{\mu}_1, \hat{\mu}_2))$ for a given number of particles $N$ ensures good sampling of the target distribution. With this automated choice of $\epsilon$, the appropriate number of $N$ should, as for the \gradient algorithm, be guided by the complexity of the performance estimate. If the performance estimate is not of sufficient quality, increasing the number of particles will decrease the minimum required trust region to sample unbiasedly from $\mu(\svec{c})$ and yield more samples for the kernel regression. It is also possible to only increase the number of particles in the buffers for the kernel regression while keeping the number of particles representing the context distribution fixed. \\
Finally, a more specific feature of \currot is the optimization of Objective (\ref{eq:currot:currot-particle-opt}), which may become more delicate in higher dimensions and require more sophisticated approaches than the simple sampling scheme used in this article. One option could be to use parallelized gradient-based optimization schemes, which should be easy to implement given the rather simple nature of the constraints. \\
\noindent \textbf{Initial Distribution in Higher Dimensions:} We also saw that using a uniform initial distribution $p_0(\svec{c})$ for \gradient and \currot can be problematic if easy tasks are unlikely under this distribution. In this case, \currot and \gradient will not achieve the expected performance threshold to progress the curriculum. Furthermore, simple search approaches for feasible contexts like the one detailed in Appendix \ref{app:currot:feasible-context-search} may fail. At this point, it may either be required to implement a more problem-specific search for feasible contexts or provide a more informed initial distribution $p_0(\svec{c})$ that does not require a search for feasible contexts. Both of these approaches can be used for \gradient and \currot.}

\end{document}